\documentclass[twoside,11pt]{article} %
\usepackage{jmlr2e}
\usepackage{times}
\usepackage{amsmath}
\usepackage{multirow}
\usepackage{booktabs}
\usepackage{hyperref,url,dsfont,nicefrac}
\hypersetup{
    colorlinks,
    breaklinks,
    linkcolor = blue,
    citecolor = blue,
    urlcolor  = blue,
}

\usepackage{algorithm,algorithmic}
\newtheorem{ass}{Assumption}
\newtheorem{thm}{Theorem}

\newtheorem{lem}{Lemma}
\usepackage{dsfont}
\def \y {\mathbf{y}}

\def \x {\mathbf{x}}

\def \w {\mathbf{w}}
\def \R {\mathbb{R}}
\def \strc {\text{sc}}
\def \strn {\textnormal{sc}}
\def \expc {\text{exp}}
\def \expn {\textnormal{exp}}
\def \con {\text{cvx}}
\def \conn {\textnormal{cvx}}

\def \P {\mathcal{P}}

\def \Y {\mathcal{Y}}

\def \A {\mathcal{A}}

\def \v {\mathbf{v}}

\def \X {\mathcal{X}}

\DeclareMathOperator*{\Reg}{Regret}

\newcommand\red[1]{{\color{red} #1}}
\def \Reg {\textsc{Reg}}
\renewcommand{\tilde}{\widetilde}
\renewcommand{\hat}{\widehat}
\allowdisplaybreaks  
\newtheorem{myRemark}{Remark}
\newcommand{\LineComment}[1]{\hfill$\rhd\ $\text{#1}}
\usepackage{algorithmic,algorithm}

\def\endenv{\hfill\raisebox{1pt}{$\triangleleft$}\smallskip}

\def \meta {\mathtt{meta}\text{-}\mathtt{regret}}
\def \expert {\mathtt{expert}\text{-}\mathtt{regret}}

\usepackage{graphicx,color} 

\usepackage{mathtools}

\begin{document}

\title{Universal Online Convex Optimization with $\mathbf{1}$ Projection per Round}

\author{\name Wenhao Yang \email yangwh@lamda.nju.edu.cn \\
       \name Yibo Wang \email wangyb@lamda.nju.edu.cn \\
       \name Peng Zhao \email zhaop@lamda.nju.edu.cn \\
       \name Lijun Zhang \email zhanglj@lamda.nju.edu.cn \\
       \addr National Key Laboratory for Novel Software Technology, Nanjing University, China\\
       \addr School of Artificial Intelligence, Nanjing University, China
       }

\maketitle

\begin{abstract}%
To address the uncertainty in function types, recent progress in online convex optimization (OCO) has spurred the development of universal algorithms that simultaneously attain minimax rates for multiple types of convex functions. However, for a $T$-round online problem, state-of-the-art methods typically conduct $O(\log T)$ projections onto the domain in each round, a process potentially time-consuming with complicated feasible sets. In this paper, inspired by the black-box reduction of \citet{pmlr-v75-cutkosky18a}, we employ a surrogate loss defined over simpler domains to develop universal OCO algorithms that only require $1$ projection. Embracing the framework of prediction with expert advice, we maintain a set of experts for each type of functions and aggregate their predictions via a meta-algorithm. The crux of our approach lies in a uniquely designed expert-loss for strongly convex functions, stemming from an innovative decomposition of the regret into the meta-regret and the expert-regret. Our analysis sheds new light on the surrogate loss, facilitating a rigorous examination of the discrepancy between the regret of the original loss and that of the surrogate loss, and carefully controlling meta-regret under the strong convexity condition. In this way, with only $1$ projection per round, we establish optimal regret bounds for general convex, exponentially concave, and strongly convex functions simultaneously. Furthermore, we enhance the expert-loss to exploit the smoothness property, and demonstrate that our algorithm can attain small-loss regret for multiple types of convex and smooth functions.  %
\end{abstract}


\section{Introduction}
Online convex optimization (OCO) stands as a pivotal online learning framework for modeling many real-world sequential predictions and decision-making problems~\citep{Intro:Online:Convex}. OCO is commonly formulated as a repeated game between the learner and the environment with the following protocol. In each round $t \in [T]$, the learner chooses a decision $\x_t$ from a convex domain $\X \subseteq \R^d$; after submitting this decision, the learner suffers a loss $f_t(\x_t)$ and observes the gradient feedback, where $f_t\colon\X\mapsto \R$ is a convex function selected by the environment. The goal of the learner is to minimize the cumulative loss over $T$ rounds, i.e., $\sum_{t=1}^T f_t(\x_t)$, and the standard performance measure is the \emph{regret}~\citep{bianchi-2006-prediction}:
\begin{equation}
  \label{eq:regret}
  \Reg_T = \sum_{t=1}^T f_t(\x_t) - \min_{\x\in\X}\sum_{t=1}^T f_t(\x),
\end{equation}
which quantifies the difference between the cumulative loss of the online learner and that of the best decision chosen in hindsight. 

Although there are plenty of algorithms to minimize the regret of convex functions, including general convex, exponentially concave (abbr.~exp-concave) and strongly convex functions~\citep{zinkevich-2003-online,ICML_Pegasos,ML:Hazan:2007}, most of them can only handle one specific function type, and need to estimate the moduli of strong convexity and exp-concavity. The demand for prior knowledge regarding function types motivates the development of \emph{universal} algorithms for OCO, which aim to attain minimax optimal regret guarantees for multiple types of convex functions simultaneously~\citep{NIPS2007_3319,NIPS2016_6268,Adaptive:Maler,pmlr-v99-mhammedi19a,ICML:2022:Zhang,NeurIPS'23:universal}.  State-of-the-art methods typically adopt a two-layer structure following the prediction with expert advice (PEA)  framework~\citep{bianchi-2006-prediction}. More specifically, they maintain $O(\log T)$ expert-algorithms with different configurations to handle the uncertainty of functions and deploy a meta-algorithm to track the best one. While this two-layer framework has demonstrated effectiveness in endowing algorithms with universality, it raises concerns regarding the computational efficiency. Since each expert-algorithm needs to execute one projection onto the feasible domain $\X$ per round, standard universal algorithms perform $O(\log T)$ projections in each round, which can be time-consuming in practical scenarios particularly when projecting onto some complicated domains. 

In the literature, there exists an effort to reduce the number of projections required by universal algorithms tailored for \emph{exp-concave functions}~\citep{pmlr-v99-mhammedi19a}. This is achieved by applying the black-box reduction of~\citep{pmlr-v75-cutkosky18a}, which reduces an OCO problem on the original (but can be complicated) feasible domain to a more manageable one on a simpler domain, such as an Euclidean ball. Deploying an existing universal algorithm~\citep{NIPS2016_6268} on the reduced problem enables us to attain optimal regret for exp-concave functions, crucially, with only \emph{one} single projection per round and no prior knowledge of exp-concavity required. However, this black-box approach cannot be extended to strongly convex functions (see Section~\ref{sec:tc} for technical discussions). Therefore, it is still unclear on how to reduce the number of projections of universal algorithms to $1$, and at the same time ensure optimal regret for strongly convex functions (as well as general convex and exp-concave functions).

\begin{table}[t]
\caption{A summary of regret of our universal algorithms and previous studies for online convex optimization over $T$ rounds $d$-dimensional functions. $L_T$ denotes the small-loss quantity. For simplicity, we use the abbreviations: cvx $\rightarrow$ convex, exp-concave $\rightarrow$ exponentially concave, str-cvx $\rightarrow$ strongly convex, \# PROJ $\rightarrow$ number of projections per round. }
\centering
\renewcommand*{\arraystretch}{1.35}
    \resizebox{0.98\textwidth}{!}{
\begin{tabular}{c|c|ccc|c}
\hline

\hline
\multirow{2}{*}{\textbf{Assumption}} & \multirow{2}{*}{\textbf{Method}} & \multicolumn{3}{c|}{\textbf{Regret Bounds}}        & \multirow{2}{*}{\textbf{\# PROJ}} \\ \cline{3-5}
  &   & cvx  & exp-concave  & str-cvx &      \\ \cline{1-6} 
 \rule{0pt}{2.5ex}    &\citet{NIPS2016_6268}  &  $O(\sqrt{T})$   &    $O(d\log T)$      &  $O(d\log T)$   &    $O(\log T)$  \\ 
 &\citet{pmlr-v99-mhammedi19a}  &  $O(\sqrt{T})$      &  $O(d\log T)$       &  $O(d\log T)$     &  $1$ \\ 
  &\citet{Adaptive:Maler} &  $O(\sqrt{T})$       &  $O(d\log T)$      &   $O(\log T)$  &  $O(\log T)$  \\ 
 & \citet{ICML:2022:Zhang} &  $O(\sqrt{T})$   &  $O(d\log T)$   &   $O(\log T)$                 &  $O(\log T)$ \\\cline{2-6} 
 & Theorem~\ref{thm:effusc} of this work &  $O(\sqrt{T})$   &  $O(d\log T)$   &   $O(\log T)$                 &  $1$ \\ \cline{1-6}
\rule{0pt}{2.5ex}  \multirow{3}{*}{$f_t(\cdot)$ is smooth} & \citet{AAAI:2020:Wang} &  $O(\sqrt{L_T})$       &  $O(d\log L_T)$      &   $O(\log L_T)$  &  $O(\log T)$ \\
 & \citet{ICML:2022:Zhang} &  $O(\sqrt{L_T})$   &  $O(d\log L_T)$   &   $O(\log L_T)$                 &  $O(\log T)$ \\ \cline{2-6}
 & Theorem~\ref{thm:effusc:smooth} of this work &  $O(\sqrt{L_T})$   &  $O(d\log L_T)$   &   $O(\log L_T)$                 &  $1$ \\
\hline

\hline
\end{tabular}}
\label{tb:results}
\end{table}

In this paper, we affirmatively solve the above question by introducing an efficient universal OCO algorithm. This algorithm necessitates only $1$ projection onto the feasible domain $\X$ per round and simultaneously delivers \emph{optimal} regret bounds for \emph{all} the three types of convex functions. Our solution employs the black-box reduction~\citep{pmlr-v119-cutkosky20a} to cast the original problem on the constrained domain $\X$ to an alternative one in terms of the domain-converting surrogate loss on a simpler domain $\Y\supseteq\X$. Specifically, we construct multiple experts updated in the domain $\Y$, each specialized for a distinct function type. Then, we combine their predictions by a meta-algorithm, and perform \emph{the only projection} onto the feasible domain $\X$. In line with previous work on universal algorithms~\citep{ICML:2022:Zhang}, the meta-algorithm chooses the linearized surrogate loss to measure the performance of experts, and is required to yield a second-order regret. The key novelty of our algorithm is the uniquely designed \emph{expert-loss for strongly convex functions}, which is motivated by an innovative decomposition of the regret into the meta-regret and expert-regret. To effectively deal with strongly convex functions, we \emph{explore the domain-converting surrogate loss in depth and illuminate its refined properties}. Our new insights tighten the regret gap in terms of original loss and surrogate loss, and further exploit strong convexity to compensate the meta-regret, thus achieving the optimal regret for strongly convex functions. Section~\ref{sec:key-ideas} provides a formal description of our key ideas. With only 1 projection per round, our algorithm attains $O(\sqrt{T})$, $O(\frac{d}{\alpha}\log T)$, and $O(\frac{1}{\lambda}\log T)$ regret for general convex, $\alpha$-exp-concave, and $\lambda$-strongly convex functions, respectively. 

We further establish the \emph{small-loss regret} for universal OCO with \emph{smooth} functions. The small-loss quantity $L_T = \min_{\x\in\X} \sum_{t=1}^T f_t(\x)$ is defined as the cumulative loss of the best decision chosen from the domain $\X$, which is at most $O(T)$ under standard OCO assumptions and meanwhile can be much smaller in benign environments. To achieve small-loss regret bounds, we design an enhanced expert-loss for smooth and strongly convex functions and integrate it into our two-layer algorithm, which finally leads to a universal OCO algorithm achieving $O(\sqrt{L_T})$, $O(\frac{d}{\alpha}\log L_T)$, and $O(\frac{1}{\lambda}\log L_T)$ small-loss regret for three types of convex functions. Notably, all those bounds are \emph{optimal} and the algorithm only requires \emph{one} projection per iteration. We summarize our results and provide a comparison to previous studies of universal algorithms in Table~\ref{tb:results}. 

\paragraph{Organization.} The rest is organized as follows. Section~\ref{sec:related-work} presents preliminaries and reviews several mostly related works. Section~\ref{sec:tcoc} illuminates  technical challenges and describes our key ideas. Section~\ref{sec:algorithms} provides the overall algorithms and regret analysis. Section~\ref{sec:ana} presents analysis of theorems and key lemmas. We finally conclude the paper in Section~\ref{sec:conclusion}. Omitted proofs and details are deferred to appendices. 

\section{Preliminaries and Related Works}
\label{sec:related-work}
In this section, we first present preliminaries, including standard assumptions of OCO, useful properties, and representative regret minimization algorithms for OCO. Then, we review several mostly related works to our paper, including universal algorithms and projection-efficient algorithms.

\subsection{Preliminaries}
We introduce two typical assumptions of online convex optimization~\citep{Intro:Online:Convex}.  
\begin{ass}[bounded domain]
\label{ass:D}
    The feasible domain $\X \subseteq \R^d$ contains the origin $\mathbf{0}$, and the diameter is bounded by $D$, i.e., $\Vert \x-\y\Vert\leq D$ holds for any $\x,\y \in \X$.
\end{ass}
\begin{ass}[bounded gradient norms]
\label{ass:G}
    The norm of the gradients of all online functions over the domain $\X$ is bounded by $G$, i.e., $\Vert \nabla f_t(\x)\Vert\leq G$ holds for all $\x \in \X$ and $t\in [T]$.
\end{ass}
Throughout the paper we use $\Vert\cdot \Vert$ for $\ell_2$-norm in default. Owing to Assumption~\ref{ass:D}, we can always construct an Euclidean ball $\Y=\{\x \mid \Vert\x\Vert\leq D\}$ containing the original feasible domain $\X$. 

Next, we state definitions of strong convexity and exp-concavity~\citep{Intro:Online:Convex}, and introduce an important property of exp-concave functions~\citep[Lemma 3]{ML:Hazan:2007}.
\begin{definition}[strongly convex functions]
\label{def:strong} 
A function $f: \X \mapsto \R$ is called $\lambda$-strongly convex, if the condition $f(\y) \geq f(\x) +  \langle \nabla f(\x), \y -\x  \rangle + \frac{\lambda}{2} \|\y -\x \|^2$ holds for all $\x,\y \in \X$.
\end{definition}
\begin{definition}[exponentially-concave functions]
\label{def:exp} 
A function $f: \X \mapsto \R$ is called $\alpha$-exponentially-concave (or, $\alpha$-exp-concave), if the function $\exp (-\alpha f(\cdot))$ is concave over the feasible domain $\X$.
\end{definition}
\begin{lem} \label{lem:exp} For an $\alpha$-exp-concave function $f:\X \mapsto \R$, if the feasible domain $\X$ has a diameter $D$ and $\|\nabla f(\x)\|\leq G$ holds for $\forall \x \in \X$, then we have
\begin{equation}\label{eqn:exp}
    f(\y) \geq f(\x)+  \langle \nabla f(\x), \y -\x  \rangle + \frac{\beta}{2}   \langle \nabla f(\x), \y -\x  \rangle^2,
\end{equation}
for all $\x,\y \in \X$, where $\beta = \frac{1}{2} \min\{\frac{1}{4GD}, \alpha \}$.
\end{lem}

There are many efforts devoted to developing regret minimization algorithms for OCO, including general convex, $\alpha$-exp-concave, and $\lambda$-strongly convex functions. For general convex functions, online gradient descent (OGD) with step size $\eta_t = O(1/\sqrt{t})$, attains an $O(\sqrt{T})$ regret~\citep{zinkevich-2003-online}. For $\alpha$-exp-concave functions, online Newton step (ONS) is equipped with an $O(\frac{d}{\alpha}\log T)$ regret~\citep{ML:Hazan:2007}. For $\lambda$-strongly convex functions, OGD with step size $\eta_t = O(1/[\lambda t])$, achieves an $O(\frac{1}{\lambda}\log T)$ regret~\citep{ICML_Pegasos}. These regret bounds are proved to be minimax optimal~\citep{Lower:bound:Portfolio,Minimax:Online}. Furthermore, problem-dependent bounds are attainable when the online functions enjoy additional properties, such as smoothness \citep{Shai:thesis,Beyond:Logarithmic,NIPS2010_Smooth,Gradual:COLT:12,Gradual:ML:14,pmlr-v40-Luo15,Adaptive:Regret:Smooth:ICML,NeurIPS'20:Sword,JMLR'24:OMD4SEA,JMLR'24:Sword++} and sparsity of gradients \citep{COLT:Adaptive:Subgradient,tieleman2012lecture,pmlr-v70-mukkamala17a,Adam,ICLR:2018:Adam,DBLP:conf/iclr/LoshchilovH19,ICLR:2020:Wang}. We discuss \emph{small-loss} regret bounds below. 

For general convex and smooth functions, \citet{NIPS2010_Smooth} prove that OGD with constant step size attains an $O(\sqrt{L})$ regret bound, where $L$ is the upper bound of $L_T$. The limitation of their method is that it requires to know $L$ beforehand. To address this issue, \citet{Adaptive:Regret:Smooth:ICML} propose scale-free online gradient descent (SOGD), which is a special case of scale-free mirror descent algorithm \citep{Scale:Free:Online}, and establish an $O(\sqrt{L_T})$ small-loss regret bound without the prior knowledge of $L_T$. For $\alpha$-exp-concave and smooth functions, ONS attains an $O(\frac{d}{\alpha} \log L_T)$ small-loss regret bound~\citep{Beyond:Logarithmic}. For $\lambda$-strongly convex and smooth functions, a variant of OGD, namely S$^2$OGD, is introduced to achieve an $O(\frac{1}{\lambda}\log L_T)$ small-loss regret bound \citep{AAAI:2020:Wang}. Such bounds reduce to the minimax optimal bounds in the worst case, but could be much tighter when the comparator has a small loss, i.e., $L_T$ is small. 

\subsection{Universal Algorithms} 
Most existing online algorithms can only handle one type of convex function and need to know the moduli of strong convexity and exp-concavity beforehand. Universal online learning aims to remove such requirements of domain knowledge. The first universal OCO algorithm is adaptive online gradient descent (AOGD)~\citep{NIPS2007_3319}, which achieves $O(\sqrt{T})$ and $O(\log T)$ regret bounds for general convex and strongly convex functions, respectively. However, the algorithm still needs to know the modulus of strong convexity and does not support exp-concave functions. 

An important milestone is the multiple eta gradient (\textsf{MetaGrad}) algorithm~\citep{NIPS2016_6268}, which can adapt to general convex and exp-concave functions without knowing the modulus of exp-concavity. MetaGrad employs a two-layer structure, which constructs multiple expert-algorithms with various learning rates and combines their predictions by a meta-algorithm called Tilted Exponentially Weighted Average (TEWA). To avoid prior knowledge of exp-concavity, each expert minimizes the expert-loss parameterized by a learning rate $\eta$, formally,
\begin{equation}\label{eq:MetaGrad:surrexp}
    \ell^{\expc}_{t,\eta} (\x)=  -\eta \langle \nabla f_t(\x_t),\x_t-\x\rangle+\eta^2 \langle \nabla f_t(\x_t),\x_t-\x\rangle^2 . 
\end{equation}
MetaGrad maintains $O(\log T)$ experts to minimize \eqref{eq:MetaGrad:surrexp}, and attains $O(\sqrt{T\log\log T})$ and $O(\frac{d}{\alpha}\log T)$ regret for general convex and $\alpha$-exp-concave functions, respectively. To further support strongly convex functions, \citet{Adaptive:Maler} propose a new type of expert-losses defined as
\begin{equation}\label{eq:Maler:surrstr}
    \ell^{\strc}_{t,\eta} (\x)=  -\eta \langle \nabla f_t(\x_t),\x_t-\x\rangle+\eta^2 G^2\Vert \x_t-\x\Vert^2  
\end{equation}
where $G$ is the gradient norm upper bound, and introduce a expert-loss for general convex functions
\begin{equation}\label{eq:Maler:surrcon}
    \ell^{\con}_{t,\eta} (\x)=  -\eta \langle \nabla f_t(\x_t),\x_t-\x\rangle+\eta^2 G^2 D^2  
\end{equation}
where $D$ is the upper bound of the diameter of $\X$. Their algorithm, named as \textsf{Maler}, obtains $O(\sqrt{T})$, $O(\frac{1}{\lambda}\log T)$ and $O(\frac{d}{\alpha}\log T)$ regret for general convex, $\lambda$-strongly convex functions, and $\alpha$-exp-concave functions, respectively. 
Later, \citet{AAAI:2020:Wang} extend Maler by replacing $G^2$ in \eqref{eq:Maler:surrstr} and \eqref{eq:Maler:surrcon} with $\Vert\nabla f_t(\x_t)\Vert^2$, thereby enabling their algorithm to deliver small-loss regret bounds. Under the smoothness condition, their algorithm  achieves $O(\sqrt{L_T})$, $O(\frac{1}{\lambda}\log L_T)$ and $O(\frac{d}{\alpha} \log L_T)$ regret for general convex, $\lambda$-strongly convex, and $\alpha$-exp-concave functions, respectively. 

MetaGrad and its variants require the carefully designed expert-losses. \citet{ICML:2022:Zhang} propose a different universal strategy that avoids the construction of losses for experts and thus can be more flexible. The basic idea is to let each expert handle original functions and deploy a meta-algorithm over \emph{linearized loss}. Importantly, the meta-algorithm is required to yield a second-order regret~\citep{pmlr-v35-gaillard14} to automatically exploit strong convexity and exp-concavity. By incorporating existing online algorithms as expert-algorithms, their approach inherits the regret of any expert designed for strongly convex functions and exp-concave functions, and also obtains minimax optimal regret (and small-loss regret) for general convex functions. 

Although state-of-the-art universal algorithms demonstrate efficacy in adapting to multiple function types, they need to create $O(\log T)$ experts to address the uncertainty of online functions. As a result, they need to perform $O(\log T)$ projections in each round, which can be time-consuming in practical scenarios with complicated feasible domains. To address this unfavorable characteristic, we aim to develop projection-efficient algorithms for universal OCO. 

\subsection{Projection-efficient Algorithms} \label{sec:pea}
In the studies of parameter-free online learning, \citet{pmlr-v75-cutkosky18a} propose a black-box reduction technique from constrained online learning to unconstrained online learning. To avoid regret degeneration, they design the \emph{domain-converting surrogate loss} $\hat{g}_t : \Y \mapsto \R$ defined as,  
\begin{equation}
\label{eq:1-projection:old}
    \hat{g}_t(\y) = \langle \nabla f_t(\x_t),\y\rangle + \Vert\nabla f_t(\x_t)\Vert \cdot S_\X(\y) 
\end{equation}
where $S_\X(\y) = \Vert \y-\Pi_\X [\y] \Vert$ is the distance function to the feasible domain $\X$. Then, we can employ an unconstrained online learning algorithm that minimizes \eqref{eq:1-projection:old} to obtain the prediction $\y_t$, and output its prediction on domain $\X$, i.e., $\x_t = \Pi_{\X}[\y_t]$. \citet[Theorem~3]{pmlr-v75-cutkosky18a} have proved that the above loss satisfies $\Vert\nabla \hat{g}_t(\y_t) \Vert\leq \Vert\nabla f_t(\x_t) \Vert$, and 
\begin{equation}\label{eqn:1-projection:old:leq}
    \langle \nabla f_t(\x_t),\x_t-\x\rangle\leq 2\big(\hat{g}_t(\y_t)-\hat{g}_t(\x) \big)\leq 2 \langle \nabla \hat{g}_t(\y_t), \y_t-\x\rangle
\end{equation}
for all $t \in [T]$ and any $\x\in\X$. Based on this fact, we know that the regret of the unconstrained problem directly serves as an upper bound for that of the original problem, hence reducing the original problem to an unconstrained surrogate problem and retaining the order of regret.

Subsequently, \citet{pmlr-v119-cutkosky20a} introduces a new domain-converting surrogate loss $g_t : \Y \mapsto \R$,
\begin{equation}\label{eq:effsurr}
    g_t(\y) = \langle \nabla f_t(\x_t), \y\rangle - \mathds{1}_{\{\langle \nabla f_t(\x_t), \v_t\rangle<0\}} \langle \nabla f_t(\x_t), \v_t\rangle \cdot S_{\X} (\y)
\end{equation}
where $\v_t = \frac{\y_t - \x_t}{\Vert \y_t - \x_t \Vert}$ is the unit vector of the projection direction. This surrogate loss enjoys more benign properties, avoiding the multiplicative constant $2$ on the right-hand side of~\eqref{eqn:1-projection:old:leq}.
\begin{lem}[{Theorem~2 of~\citet{pmlr-v119-cutkosky20a}}]
\label{thm:effsurr}
The function defined in \eqref{eq:effsurr} is convex, and it satisfies $\Vert\nabla g_t(\y_t) \Vert\leq \Vert\nabla f_t(\x_t) \Vert$.
Furthermore, for all $t$ and all $\x\in\X$, we have
    \begin{equation}\label{eqn:effsurr:leq}
        \langle \nabla f_t(\x_t),\x_t-\x\rangle\leq g_t(\y_t)-g_t(\x)  \leq \langle \nabla g_t(\y_t),\y_t-\x\rangle. 
    \end{equation}
\end{lem}

While the black-box reduction is initially proposed for the constrained-to-unconstrained conversion, it also facilitates the conversion to another constrained problem (i.e., $\Y \neq \R^d$). This enables us to transform OCO problem on a complicated domain into another on simpler domains such that the projection is much easier. Building on this idea, \citet{pmlr-v99-mhammedi19a} introduce an efficient implementation of MetaGrad \citep{NIPS2016_6268}, which only conducts $1$ projection onto the original domain in each round, and keeps the same order of regret bounds. However, as detailed in the following section, the black-box reduction does not adequately extend to strongly convex functions. We also mention that~\citet{zhaoefficient} recently employ the technique to non-stationary OCO with non-trivial modifications to develop efficient algorithms for minimizing dynamic regret and adaptive regret. However, they focus on the convex functions and do not involve the considerations of exp-concave and strongly convex functions as concerned in our paper.

\section{Technical Challenge and Our Key Ideas}
\label{sec:tcoc}
In this section, we elaborate on the technical challenges and our key ideas. 

\subsection{Technical Challenge}
\label{sec:tc}
As mentioned,~\citet{pmlr-v99-mhammedi19a} exploit the black-box reduction scheme of~\citet{pmlr-v75-cutkosky18a} to improve the projection efficiency of MetaGrad~\citep{NIPS2016_6268}. We summarize their algorithm in Algorithm~\ref{alg:effMetaGrad}. In the following, we will demonstrate its effectiveness for exp-concave functions and explain why it fails for strongly convex functions. 

\paragraph{Success in Exp-concave Functions.} By applying the black-box reduction as described in Section~\ref{sec:pea}, \citet{pmlr-v99-mhammedi19a} utilize MetaGrad to minimize the surrogate loss $\hat{g}_t(\cdot)$ in \eqref{eq:1-projection:old} over an Euclidean ball $\Y$. The projection operations inside MetaGrad are over $\Y$ and thus negligible. Notice that Algorithm~\ref{alg:effMetaGrad} demands only $1$ projection onto $\X$ in Step~4. According to regret bound of MetaGrad, Algorithm~\ref{alg:effMetaGrad} enjoys a second-order bound~\citep[Theorem~10]{pmlr-v99-mhammedi19a},  
\begin{equation}\label{eqn:MetaGrad:second-order}
    \sum_{t=1}^T \langle \nabla \hat{g}_t(\y_t),\y_t-\x\rangle \leq O\left( \sqrt{d\log T \cdot \sum_{t=1}^T \langle \nabla \hat{g}_t(\y_t),\y_t-\x\rangle^2} + d\log T\right). 
\end{equation}
The above bound is measured in terms of the surrogate loss, thus requiring a further analysis that converts it back to the bound of the original function. Since $\beta = \frac{1}{2}\min \left\{ \frac{1}{4GD},\alpha \right\}$, the function $x-\beta x^2$ is strictly increasing when $x\in  (-\infty, 2GD]$. 
Therefore, the property of the domain-converting surrogate loss $\hat{g}_t(\cdot)$ in \eqref{eqn:1-projection:old:leq} implies 
\begin{equation}\label{eqn:MetaGrad:leq}
    \frac{1}{2}\langle \nabla f_t(\x_t),\x_t-\x\rangle -\frac{\beta}{4}\langle \nabla f_t(\x_t),\x_t-\x\rangle^2  \leq \langle \nabla \hat{g}_t(\y_t),\y_t-\x\rangle-\beta \langle \nabla \hat{g}_t(\y_t),\y_t-\x\rangle^2. 
\end{equation}
Combining \eqref{eqn:MetaGrad:second-order} with \eqref{eqn:MetaGrad:leq} and applying the AM-GM inequality, we obtain 
\begin{equation*}
\sum_{t=1}^T\langle \nabla f_t(\x_t),\x_t-\x\rangle -\frac{\beta}{2}\sum_{t=1}^T\langle \nabla f_t(\x_t),\x_t-\x\rangle^2\leq O\left(\frac{d}{\alpha}\log T\right), 
\end{equation*}
thereby achieving the optimal regret for $\alpha$-exp-concave functions based on Lemma~\ref{lem:exp}. 

\begin{algorithm}[t]
\caption{{Black-box reduction for projection-efficient MetaGrad~\citep{pmlr-v99-mhammedi19a}}}
\label{alg:effMetaGrad}
\begin{algorithmic}[1]
   \STATE Construct a ball domain $\Y=\{\x \mid \Vert\x\Vert\leq D\}\supseteq\X$
   \FOR{$t=1$ \TO $T$} 
   \STATE Receive the decision $\y_t\in\Y$ from MetaGrad 
   \STATE Submit the decision $\x_t= \Pi_\X [\y_t]$ \LineComment{The only step projects onto domain $\X$ per round.} 
   \STATE Suffer the loss $f_t(\x_t)$ and observe the gradient $\nabla f_t(\x_t)$
   \STATE Construct the surrogate loss $\hat{g}_t(\cdot)$ as~\eqref{eq:1-projection:old} and send it to MetaGrad
   \ENDFOR
\end{algorithmic}
\end{algorithm}

\paragraph{Failure in Strongly Convex Functions.} 
To handle strongly convex functions, a straightforward way is to use a universal algorithm that supports strongly convex functions, such as Maler~\citep{Adaptive:Maler}, as the black-box subroutine in Algorithm~\ref{alg:effMetaGrad}. However, for strongly convex functions, the above analysis cannot be applied, and we are unable to derive a tight regret bound. 
Specifically, according to the theoretical guarantee of Maler \citep[Theorem~1]{Adaptive:Maler}, we have
\begin{equation}\label{eqn:Maler}
    \sum_{t=1}^T \langle \nabla \hat{g}_t(\y_t),\y_t-\x\rangle \leq O\left( \sqrt{\log T \cdot \sum_{t=1}^T \Vert \y_t-\x\Vert^2} + \log T\right). 
\end{equation}
From the standard black-box analysis and the definition  of strong convexity, we know
\begin{equation}\label{eqn:str:standard}
    \begin{aligned}
   \sum_{t=1}^T f_t(\x_t) - \sum_{t=1}^T f_t(\x)  
       \overset{\eqref{eqn:1-projection:old:leq}}{\leq} 2\sum_{t=1}^T \langle \nabla \hat{g}_t(\y_t),\y_t-\x\rangle - \frac{\lambda}{2} \sum_{t=1}^T \Vert \x_t-\x\Vert^2.
    \end{aligned}
\end{equation}
Substituting \eqref{eqn:Maler} into \eqref{eqn:str:standard}, we 
 encounter an $\tilde{O}(\sqrt{\sum_{t=1}^T \Vert \y_t-\x\Vert^2} - \frac{\lambda}{2}\sum_{t=1}^T \Vert \x_t-\x\Vert^2)$ term, which is hard to manage due to $\Vert \y_t-\x\Vert\geq \Vert \x_t-\x\Vert$. Here, $\tilde{O}(\cdot)$ omits the $\text{ploy}(\log T)$ factors.  

\subsection{Key Ideas} 
\label{sec:key-ideas}
To address above challenges, we introduce novel ideas in both algorithm design and regret analysis. 

\paragraph{Algorithm Design.} Our algorithm is still in a two-layer structure. The main contribution lies in a uniquely designed \emph{expert-loss for strongly convex functions}. For simplicity, we consider that the modulus of strong convexity $\lambda$ is known for a moment, and define  
\begin{equation}
\label{eq:effsurr:str}
    \ell^{\strc}_{t} (\y)= \langle \nabla g_t(\y_t),\y\rangle + \frac{\lambda}{2} \Vert \y-\red{\x_t}\Vert^2, 
\end{equation}
where $g_t(\cdot)$ is the surrogate loss defined in~\eqref{eq:effsurr}. Next, we shall  compare our designed expert-loss~\eqref{eq:effsurr:str} with the one when applying existing universal algorithms in a black-box manner. Suppose Maler~\citep{Adaptive:Maler} is used, their expert-loss construction~\eqref{eq:Maler:surrstr} indicates that the algorithm within $\Y$ domain essentially optimizes the following expert-loss (up to constant factors):
\begin{equation}
\label{eq:effsurr:str-opposed}
    \hat{\ell}^{\strc}_{t} (\y)= \langle \nabla g_t(\y_t),\y\rangle + \frac{\lambda}{2} \Vert \y-\y_t\Vert^2. 
\end{equation}
An important caveat in our approach is that our expert-loss evaluates the performance of the expert (associated with strongly convex functions) based on the distance between its output $\y$ and the \emph{actual} decision $\x_t\in \X$, as opposed to the unprojected intermediate one $\y_t\in \Y$ in~\eqref{eq:effsurr:str-opposed}. 

In fact, this design of expert-loss~\eqref{eq:effsurr:str} stems from a novel regret decomposition as explained below. First, by strong convexity of $f_t$, we have
\begin{equation}
\label{eqn:decom:orig:old}
    \begin{aligned}
        & \sum_{t=1}^T f_t(\x_t) - \sum_{t=1}^T f_t(\x) \leq \sum_{t=1}^T \langle\nabla f_t(\x_t),\x_t-\x \rangle - \frac{\lambda}{2} \sum_{t=1}^T \Vert \x_t-\x\Vert^2  \\
        \overset{\eqref{eqn:effsurr:leq}}{\leq} {} & \sum_{t=1}^T \langle \nabla g_t(\y_t),\y_t-\x\rangle - \frac{\lambda}{2} \sum_{t=1}^T \Vert \x_t-\x\Vert^2  \\
        = {} &  \underbrace{\sum_{t=1}^T \langle \nabla g_t(\y_t),\y_t-\y^i_t\rangle}_{\meta} + \sum_{t=1}^T\langle \nabla g_t(\y_t),\y_t^i-\x \rangle - \frac{\lambda}{2} \sum_{t=1}^T \Vert \x_t-\x\Vert^2,  
\end{aligned}
\end{equation}
where $\y_t^i$ denotes the decision of the $i$-th expert. The first term of the above bound is the meta-regret in terms of linearized surrogate loss. Then, we reformulate the remaining two terms as follows
\begin{equation}
    \label{eq:effstrdecom}
    \begin{aligned}
        &\sum_{t=1}^T\langle \nabla g_t(\y_t),\y_t^i-\x \rangle - \frac{\lambda}{2} \sum_{t=1}^T \Vert \x_t-\x\Vert^2    
           =  \sum_{t=1}^T \left( \langle \nabla g_t(\y_t),\y_t^i \rangle + \frac{\lambda}{2} \Vert \x_t-\y_t^i\Vert^2 \right)     \\
        & - \sum_{t=1}^T \left( \langle \nabla g_t(\y_t),\x \rangle + \frac{\lambda}{2} \Vert \x_t-\x\Vert^2 \right) - \frac{\lambda}{2} \sum_{t=1}^T \Vert \x_t-\y^i_t\Vert^2  \\
        \overset{\eqref{eq:effsurr:str}}{=} {} & \underbrace{\sum_{t=1}^T \Big( \ell^{\strc}_t (\y_t^i) - \ell^{\strc}_t (\x) \Big)}_{\expert} - \frac{\lambda}{2} \sum_{t=1}^T \Vert \x_t-\y^i_t\Vert^2,
    \end{aligned}
\end{equation}
where the expert-loss in \eqref{eq:effsurr:str} naturally arises. Combining \eqref{eqn:decom:orig:old} with \eqref{eq:effstrdecom}, we arrive at
\begin{equation}\label{eqn:decom:orig}
    \begin{aligned}
         \sum_{t=1}^T f_t(\x_t) - \sum_{t=1}^T f_t(\x)  
         \leq   \underbrace{\sum_{t=1}^T \langle \nabla g_t(\y_t),\y_t-\y^i_t\rangle}_{\meta} + \underbrace{\sum_{t=1}^T \Big( \ell^{\strc}_t (\y_t^i) - \ell^{\strc}_t (\x) \Big)}_{\expert} - \frac{\lambda}{2} \sum_{t=1}^T \Vert \x_t-\y^i_t\Vert^2. 
    \end{aligned}
\end{equation}

\paragraph{Theoretical Analysis.} For the expert-regret, since expert-loss~\eqref{eq:effsurr:str} is $\lambda$-strongly convex and its gradients are bounded (see Lemma~\ref{lem:surrstr}), we can directly use OGD to achieve an optimal $O(\frac{1}{\lambda}\log T)$ regret. Thus, we proceed to handle the meta-regret. In line with the research of universal algorithms~\citep{ICML:2022:Zhang}, we require the meta-algorithm to yield a second-order regret bound 
\begin{equation}\label{eqn:meta:second-order}
    \sum_{t=1}^T \langle \nabla g_t(\y_t),\y_t-\y^i_t\rangle \leq O\left( \sqrt{\sum_{t=1}^T \langle \nabla g_t(\y_t),\y_t-\y^i_t\rangle^2} \right).  
\end{equation} 
Notably, the upper bound of \eqref{eqn:meta:second-order} and the negative term in \eqref{eqn:decom:orig} cannot be canceled due to the dismatch between $\y_t-\y^i_t$ and $\x_t-\y^i_t$. To resolve this discrepancy, we demonstrate that the surrogate loss defined in~\eqref{eq:effsurr} enjoys the following two important improved properties. 
\begin{lem}
\label{lem:effsurr}
In addition to enjoying all the properties outlined in Lemma~\ref{thm:effsurr}, the surrogate loss function $g_t:\Y \mapsto \R$ defined in \eqref{eq:effsurr} satisfies
\begin{equation}\label{eqn:effsur:leqimprove}
        \langle \nabla f_t(\x_t),\x_t-\x \rangle \leq \langle \nabla g_t(\y_t),\y_t-\x\rangle \red{- \mathds{1}_{\{ \langle\nabla f_t(\x_t),\v_t\rangle\geq 0 \}} \cdot \langle\nabla f_t(\x_t),\y_t-\x_t \rangle},
\end{equation}
for all $t$ and all $\x\in\X$. Furthermore, we also have 
\begin{equation}\label{eqn:con:xtyt}
\left\{
\begin{array}{ll}
\langle \nabla g_t(\y_t), \x_t-\y_t\rangle=0, & \text{when } \langle \nabla f_t(\x_t),\v_t\rangle<0, \vspace{2mm}\\
\langle \nabla g_t(\y_t), \x_t-\y_t\rangle\leq 0, & \text{otherwise}.  
\end{array} \right.
\end{equation}
\end{lem}

\begin{myRemark}\normalfont
We highlight the improvements of Lemma~\ref{lem:effsurr} over Lemma~\ref{thm:effsurr}. First, we provide a tighter connection between the linearized online function and the surrogate loss in \eqref{eqn:effsur:leqimprove}. Second, we analyze the difference between the actual decision $\x_t$ and the intermediate decision $\y_t$, along the direction $\nabla g_t(\y_t)$ in \eqref{eqn:con:xtyt}. As shown later, both of them are crucial for controlling the meta-regret.
\endenv
\end{myRemark}

Utilizing \eqref{eqn:effsur:leqimprove} in Lemma~\ref{lem:effsurr}, we refine the decomposition in \eqref{eqn:decom:orig} to establish a tighter bound  
\begin{equation}\label{eqn:decom}
    \sum_{t=1}^T f_t(\x_t) - \sum_{t=1}^T f_t(\x)  \overset{\eqref{eqn:decom:orig:old},\eqref{eq:effstrdecom},\eqref{eqn:effsur:leqimprove}}{\leq}  \sum_{t=1}^T \langle \nabla g_t(\y_t),\y_t-\y^i_t\rangle + \text{ER}(T) - \frac{\lambda}{2} \sum_{t=1}^T \Vert \x_t -\y_t^i \Vert^2 - \red{\Delta_T}
\end{equation}
where $\text{ER}(T) =\sum_{t=1}^T \ell^{\strc}_t (\y_t^i) - \sum_{t=1}^T \ell^{\strc}_t (\x) =O(\frac{1}{\lambda}\log T)$ is the expert-regret and $\Delta_T = \sum_{t=1}^T \mathds{1}_{\{ \langle\nabla f_t(\x_t),\v_t\rangle\geq 0 \}} \cdot \langle\nabla f_t(\x_t),\y_t-\x_t \rangle\geq 0$ is the crucial negative term introduced in the surrogate loss. Compared to \eqref{eqn:decom:orig}, the new upper bound~\eqref{eqn:decom} enjoys an additional negative term $-\Delta_T$, which is essential to achieve a favorable regret bound in the analysis.  

To utilize the negative quadratic term $-\frac{\lambda}{2}\sum_{t=1}^T \Vert \x_t -\y_t^i \Vert^2$ in \eqref{eqn:decom} for compensating the second-order bound in \eqref{eqn:meta:second-order}, we need to convert $\y_t$ to $\x_t$, a place where  \eqref{eqn:con:xtyt} comes into play. From \eqref{eqn:meta:second-order} and \eqref{eqn:con:xtyt}, we prove that for any $\gamma\in (0,\frac{G}{2D}]$ it holds that (see Lemma~\ref{lem:meta-regret:effsurr} for details): 
\begin{equation}\label{eqn:meta-regret}
    \sum_{t=1}^T \langle \nabla g_t(\y_t),\y_t-\y^i_t\rangle 
     \leq O\left( \frac{G^2}{2\gamma} \right) + \frac{\gamma}{2G^2}\sum_{t=1}^T \langle \nabla g_t(\y_t),\x_t-\y^i_t\rangle^2 + \red{\Delta_T}. 
\end{equation} 
Substituting \eqref{eqn:meta-regret} into \eqref{eqn:decom}, the additional term $\Delta_T$ is automatically \textit{canceled out}, and we have
\begin{equation*}
    \begin{aligned}
        \sum_{t=1}^T f_t(\x_t) - \sum_{t=1}^T f_t(\x)  
        &\leq  \text{ER}(T)+ O\left( \frac{G^2}{2\gamma} \right) + \frac{\gamma}{2G^2}\sum_{t=1}^T \langle \nabla g_t(\y_t),\x_t-\y^i_t\rangle^2 - \frac{\lambda}{2} \sum_{t=1}^T \Vert \x_t -\y_t^i \Vert^2 \\
        &\leq \text{ER}(T)+ O\left( \frac{G^2}{2\gamma} \right) + \frac{\gamma}{2}\sum_{t=1}^T \Vert\x_t-\y^i_t\Vert^2 - \frac{\lambda}{2} \sum_{t=1}^T \Vert \x_t -\y_t^i \Vert^2 \\
        &\leq O\left( \frac{G^2}{2\gamma}\right) + \text{ER}(T) = O\left( \frac{1}{\lambda} \log T\right)
    \end{aligned}
\end{equation*}
where the last inequality is because we set $\gamma = \min \{\frac{G}{2D},\lambda\}$.

\begin{myRemark}\normalfont
Section~\ref{sec:pea} describes two kinds of surrogate loss developed in parameter-free online learning, as specified in~\eqref{eq:1-projection:old} and~\eqref{eq:effsurr}. Indeed, they \emph{both} are suitable for parameter-free online learning~\citep{pmlr-v119-cutkosky20a} and reducing projection complexity for non-stationary online learning~\citep{zhaoefficient}, with the new one offering an improvement in terms of a multiplicative constant $2$. However, it is essential to adopt the new surrogate loss in our purpose: as established in Lemma~\ref{lem:effsurr}, both negative terms and the mild difference between $\x_t$ and $\y_t$ play a critical role in our regret analysis. By contrast, the old surrogate loss~\eqref{eq:1-projection:old} lacks these advanced properties.
\endenv
\end{myRemark}

\section{Efficient Algorithm for Universal Online Convex Optimization}
\label{sec:algorithms}
In this section, we provide the details of our developed efficient algorithms for universal OCO, following the key ideas presented in Section~\ref{sec:key-ideas}. We construct a set of experts for each type of functions and use a meta-algorithm to combine their predictions. To reduce the cost of projections, these experts are updated on an Euclidean ball $\Y = \{\x \mid \Vert\x\Vert\leq D\}$ enclosing the feasible domain $\X$. After combining their decisions via the meta-algorithm, we project the solution in $\Y$ onto domain $\X$, which is the only projection onto $\X$ per round. 
\begin{algorithm}[t]
   \caption{Efficient Algorithm for Universal OCO}\label{alg:effusc}
\begin{algorithmic}[1]
   \STATE \textbf{Input:} The modulus set $\P_{\strc}$ and $\P_{\expc}$, the expert set $\A=\emptyset$, the number of experts $k=0$
   \STATE $k \leftarrow k+1$, create a expert $E^1$ by running OGD with loss~\eqref{eq:effUSC:surr:con} over $\Y$
   \FORALL{$\hat{\alpha}\in\P_{\expc}$}
   \STATE $k \leftarrow k+1$, create a expert $E^k$ by running ONS  with loss~\eqref{eq:effUSC:surr:exp} and parameter $\hat{\alpha}$ over $\Y$
   \ENDFOR
   \FORALL{$\hat{\lambda}\in\P_{\strc}$}
   \STATE $k \leftarrow k+1$, create a expert $E^k$ by running OGD with loss~\eqref{eq:effUSC:surr:str} and parameter $\hat{\lambda}$ over $\Y$
   \ENDFOR
   \STATE Add all the experts to the set: $\A= \{ E^1,E^2,\cdots,E^k \}$
   \FOR{$t=1$ \TO $T$} 
   \STATE Compute the weight $p^i_{t}$ of each expert $E^i$ by \eqref{eq:effUSC:pti}
   \STATE Receive the decision $\y^i_{t}$ from each expert $E^i$ in $\A$ 
   \STATE Aggregate all the decisions by $\y_t = \sum_{i=1}^{|\A|} p_t^i \y_t^i$ 
   \STATE Submit the decision $\x_t= \Pi_\X [\y_t]$ \LineComment{The only step projects onto domain $\X$ per round.} 
   \STATE Suffer the loss $f_t(\x_t)$ and observe the gradient $\nabla f_t(\x_t)$
   \STATE Construct the expert-loss $\ell^{\con}_t(\cdot)$, $\ell^{\strc}_t(\cdot)$ or $\ell^{\expc}_t(\cdot)$ and send it to corresponding expert in $\A$
   \ENDFOR
\end{algorithmic}
\end{algorithm}

\subsection{Efficient Algorithm for Minimax Universal Regret}
To handle unknown parameters of strong convexity and exp-concavity, we construct two finite sets, i.e., $\P_{\strc}$ and $\P_{\expc}$, to approximate their values. Taking $\lambda$-strongly convex functions as an example, we assume the unknown modulus $\lambda$ is bounded by $\lambda\in [1/T,1]$\footnote{One can verify the degenerated situations where the unknown modulus falls outside the range, which will not be a concern. Formal justifications are provided in Appendix~\ref{app:boundmodulus}. \label{footnote:AppD}}, and set $\P_{\strc}=\{ 1/T,2/T,\cdots,2^N/T \}$, where $N=\lceil \log_2 T\rceil$. In this way, for any $\lambda\in [1/T,1]$, there exists a $\hat{\lambda}\in \P_{\strc}$ such that $\hat{\lambda}\leq \lambda \leq 2\hat{\lambda}$. Moreover, we design three types of expert-losses. For general convex functions, it is defined as
\begin{equation}\label{eq:effUSC:surr:con}
     \ell^{\con}_t (\y)= \langle \nabla g_t(\y_t),\y-\y_t\rangle, 
 \end{equation}
 where $g_t(\y)$ is defined in \eqref{eq:effsurr}. We can then use OGD as the expert-algorithm to minimize the regret. To handle exp-concave functions, we construct the expert-loss for each $\hat{\alpha}\in \P_{\expc}$ as
\begin{equation}\label{eq:effUSC:surr:exp}
     \ell^{\expc}_{t,\hat{\alpha}} (\y)= \langle \nabla g_t(\y_t),\y-\y_t\rangle + \frac{\hat{\beta}}{2} \langle \nabla g_t(\y_t),\y-\y_t\rangle^2,  
 \end{equation}
 where $\hat{\beta} = \frac{1}{2} \min\{\frac{1}{4GD}, \hat{\alpha} \}$. It is easy to verify that  $\ell^{\expc}_{t,\hat{\alpha}}(\cdot)$ is $\frac{\hat{\beta}}{4}$-exp-concave, so we use ONS  as the expert-algorithm. To handle strongly convex functions, as discussed in Section~\ref{sec:key-ideas}, we construct the following  expert-loss for each $\hat{\lambda}\in \P_{\strc}$ whose quadratic proximal regularizer is using $\x_t$,
\begin{equation}\label{eq:effUSC:surr:str}
     \ell^{\strc}_{t,\hat{\lambda}} (\y)= \langle \nabla g_t(\y_t),\y-\y_t\rangle + \frac{\hat{\lambda}}{2} \Vert \y-\x_t\Vert^2.  
 \end{equation}
 Since $\ell^{\strc}_{t,\hat{\lambda}}(\cdot)$ is $\hat{\lambda}$-strongly convex, we use OGD with step size $\eta_t=1/[\hat{\lambda}t]$ as the expert-algorithm. 
Finally, we deploy a meta-algorithm to track the best expert on the fly. Following \citet{ICML:2022:Zhang}, we use the linearized surrogate loss to measure the performance of experts, and choose Adapt-ML-Prod \citep{pmlr-v35-gaillard14} as the meta-algorithm to yield a second-order bound. 

Our efficient algorithm for universal OCO is summarized in Algorithm~\ref{alg:effusc}. From Steps~2 to 9, it creates a set of experts by running multiple  algorithms over the ball $\Y$, each specialized for a distinct function type. Then, it maintains a set $\A$ consisting of all the experts, and the $i$-th expert is denoted by $E^i$. In the $t$-th round, it computes the weight $p_t^i$ of each expert $E^i$ in Step~11 according to Adapt-ML-Prod.  After receiving all the predictions in Step~12, it aggregates them based on their weights to attain $\y_t$ in Step~13. Next, it conducts the \textit{only} projection onto the original domain $\X$ to obtain the actual decision $\x_t$ in Step~14. In Step~15, it evaluates the gradient $\nabla f_t(\x_t)$ to construct the expert-losses in \eqref{eq:effUSC:surr:con}, \eqref{eq:effUSC:surr:exp}, and \eqref{eq:effUSC:surr:str}. In Step~16, it sends the corresponding expert-loss to each expert so that it can make predictions for the next round. 

Finally, we elucidate how our algorithm determines the weight of the $i$-th expert $E^i$. We measure the performance of expert $E^i$ by the linearized surrogate loss, i.e., $l_t^i = \langle \nabla g_t(\y_t), \y_t^i - \y_t\rangle$. According to Lemma~\ref{thm:effsurr}, we have $|l_t^i|\leq \Vert \nabla g_t(\y_t) \Vert \Vert \y_t^i - \y_t\Vert \leq 2GD$. Since \mbox{Adapt-ML-Prod} requires the loss to fall within the range of $[0,1]$, we normalize $l_t^i$ to construct the meta-loss as
$\ell_t^i = (\langle \nabla g_t(\y_t), \y_t^i - \y_t\rangle)/(4GD)+\frac{1}{2}\in [0,1]$.  The loss of the meta-algorithm in the $t$-th round is  $\ell_t = \sum_{i=1}^{|\A|} p_t^i \ell_t^i$, which is a constant $\frac{1}{2}$ due to its construction and Step~13. For the expert $E^i$, its weight is updated by \mbox{Adapt-ML-Prod} algorithm~\citep{pmlr-v35-gaillard14} in the following way:
 \begin{equation}\label{eq:effUSC:pti}
     p_t^i = \frac{\eta_{t-1}^i w_{t-1}^i}{\sum_{j=1}^{|\A|} \eta_{t-1}^j w_{t-1}^j},~~w^i_{t-1} = \left( w^i_{t-2} \left( 1+\eta^i_{t-2} (\ell_{t-1}-\ell^i_{t-1}) \right) 
     \right)^{\frac{\eta^i_{t-1}}{\eta^i_{t-2}}}
 \end{equation}
 where $\eta^i_{t-1} = \min \big\{ \frac{1}{2},\sqrt{(\ln |\A|)/(1+\sum_{s=1}^{t-1} (\ell_s -\ell_s^i)^2)} \big\}$. In the first round, we set $w_0^i=1/|\A|$.

\begin{myRemark}\normalfont 
While the surrogate loss \eqref{eq:effsurr} involves the projection operation, our proposed meta-loss and expert-losses only access the gradient $g_t(\y)$ through $\nabla g_t(\y_t)$, which is given by~\citet{pmlr-v119-cutkosky20a}, 
\begin{equation*}
    \nabla g_t(\y_t) = \nabla f_t(\x_t) - \mathds{1}_{\{ \langle \nabla f_t(\x_t),\v_t \rangle < 0 \}} \langle \nabla f_t(\x_t),\v_t \rangle \cdot \v_t
\end{equation*}
 where $\v_t = \frac{\y_t - \x_t}{\Vert \y_t - \x_t \Vert}$. According to the above formulation, the gradient can be directly computed from $\x_t$ and $\y_t$, which means \emph{no} additional projections are needed at each round. Therefore, our algorithm requires only $1$ projection onto domain $\X$ per round.
\endenv
\end{myRemark}

Below, we provide the meta-regret analysis of Algorithm~\ref{alg:effusc}, and defer the details of expert-algorithms and related analysis in Appendix~\ref{app:expert}.
\begin{lem}\label{lem:meta-regret:effsurr}
Under Assumptions~\ref{ass:D} and \ref{ass:G}, the meta-regret of Algorithm~\ref{alg:effusc} satisfies
\begin{equation*}
    \begin{aligned}
        & \sum_{t=1}^T \langle \nabla g_t(\y_t),\y_t-\y_t^i \rangle \leq 8\Gamma GD + \frac{\Gamma}{\sqrt{\ln |\A|}} \sqrt{16G^2 D^2 + \sum_{t=1}^T \langle \nabla g_t(\y_t),\red{\y_t}-\y_t^i \rangle^2 } \\
        \leq {}& 4\Gamma GD \left( 2+\frac{1}{\sqrt{\ln |\A|}} \right) + \frac{\Gamma^2 G^2}{2\gamma \ln |\A|} + \frac{\gamma}{2G^2} \sum_{t=1}^T \langle \nabla g_t(\y_t),\red{\x_t}-\y_t^i \rangle^2 +\red{\Delta_T}
    \end{aligned}
\end{equation*}
for any $\gamma\in (0,\frac{G}{2D}]$, where $\Delta_T = \sum_{t=1}^T \mathds{1}_{\{ \langle\nabla f_t(\x_t),\v_t\rangle\geq 0 \}} \cdot \langle\nabla f_t(\x_t),\y_t-\x_t \rangle$, and $\Gamma = 3\ln |\A|+\ln (1+\frac{|\A|}{2e}(1+\ln(T+1)))=O(\log\log T)$. 
\end{lem}
As mentioned in Section~\ref{sec:key-ideas}, Lemma~\ref{lem:meta-regret:effsurr} is pivotal in dealing with technical challenge. Specifically, when the meta-algorithm enjoys a second-order bound in terms of the surrogate loss in \eqref{eq:effsurr}, we can then convert the intermediate decision $\y_t$ in the meta-regret bound to the actual one $\x_t$ at the cost of adding an addition positive term, as presented in the analysis in~\eqref{eqn:meta-regret}.

Based on Lemma~\ref{lem:meta-regret:effsurr}, we present the following theoretical guarantee of Algorithm~\ref{alg:effusc}.
\begin{thm}\label{thm:effusc}
    Under Assumptions~\ref{ass:D} and \ref{ass:G}, Algorithm~\ref{alg:effusc} attains $O(\sqrt{T})$, $O(\frac{d}{\alpha}\log T)$ and $O(\frac{1}{\lambda}\log T)$ regret for general convex functions, $\alpha$-exp-concave functions with $\alpha\in[1/T,1]$, and $\lambda$-strongly convex functions with $\lambda\in[1/T,1]$, respectively. Moreover, Algorithm~\ref{alg:effusc} requires only $1$ projection onto the feasible domain $\X$ per round. 
\end{thm}
\begin{myRemark}\normalfont 
Similar to previous studies \citep{Adaptive:Maler,ICML:2022:Zhang}, our universal algorithm also achieves the minimax optimal regret, but only requires $1$ projection.
\endenv
\end{myRemark}

\subsection{Efficient Algorithm for Small-Loss Universal Regret}
Furthermore, we consider the small-loss regret for smooth and non-negative online functions. To this end, an additional assumption is required~\citep{NIPS2010_Smooth}
\begin{ass}\label{ass:H}
    All the online functions are non-negative, and $H$-smooth over $\X$.  \footnote{For simplicity, we require the online functions to be non-negative, otherwise, one may redefine the small-loss quantity as $L_T = \min_{\x \in \X} \sum_{t=1}^T f_t(\x) - \sum_{t=1}^T\min_{\x \in \X}  f_t(\x)$ as suggested in~\citep[Theorem 4.23]{Modern:Online:Learning}.}
\end{ass}
To exploit the smoothness, we enhance the expert-loss for strongly convex functions in \eqref{eq:effUSC:surr:str} as 
\begin{equation}\label{eq:effUSC:surr:str:small-loss}
     \hat{\ell}^{\strc}_{t,\hat{\lambda}} (\y)= \langle \nabla g_t(\y_t),\y-\y_t\rangle + \frac{\hat{\lambda}}{2G^2} \Vert  \nabla g_t(\y_t) \Vert^2\Vert \y-\x_t\Vert^2.  
 \end{equation}
 Since $\hat{\ell}^{\strc}_{t,\hat{\lambda}} (\cdot)$ is $\frac{\hat{\lambda}}{G^2}\Vert  \nabla g_t(\y_t) \Vert^2$-strongly convex and smooth, we use S$^2$OGD \citep{AAAI:2020:Wang} as the expert-algorithm to deliver small-loss expert-regret. 
For general convex and exp-concave functions, we reuse~\eqref{eq:effUSC:surr:con} and~\eqref{eq:effUSC:surr:exp}  as the expert-losses, and employ ONS~\citep{Beyond:Logarithmic} and SOGD~\citep{Adaptive:Regret:Smooth:ICML} as the expert-algorithms to deliver small-loss expert-regret. The meta-algorithm remains unchanged. In this way, we get the following regret guarantee.
 \begin{thm}\label{thm:effusc:smooth}
    Under Assumptions~\ref{ass:D}, \ref{ass:G} and \ref{ass:H}, the improved version of Algorithm~\ref{alg:effusc} attains $O(\sqrt{L_T})$, $O(\frac{d}{\alpha}\log L_T)$ and $O(\frac{1}{\lambda}\log L_T)$ regret for general convex functions, $\alpha$-exp-concave functions with $\alpha\in[1/T,1]$, and $\lambda$-strongly convex functions with $\lambda\in[1/T,1]$, respectively, where the small-loss quantity $L_T = \min_{\x\in\X} \sum_{t=1}^T f_t(\x)$ is the cumulative loss of the best decision from the domain $\X$. Moreover, the overall algorithm requires only $1$ projection onto the feasible domain $\X$ per round. 
\end{thm}
\begin{myRemark}\normalfont 
With only $1$ projection in each round, our universal algorithm is able to deliver \emph{optimal} small-loss regret bounds for multiple types of convex functions simultaneously. In contrast, \citet{AAAI:2020:Wang} and \citet{ICML:2022:Zhang} take $O(\log T)$ projections to achieve the small-loss regret.
\endenv
\end{myRemark}

\section{Analysis}\label{sec:ana}
We prove Lemma~\ref{lem:effsurr}, Lemma~\ref{lem:meta-regret:effsurr}, Theorem~\ref{thm:effusc}, and Theorem~\ref{thm:effusc:smooth} in this section. The proofs of supporting lemmas can be found in the Appendix~\ref{app:supporting}.

\subsection{Proof of Lemma~\ref{lem:effsurr}}
 According to \eqref{eq:effsurr}, the (sub-)gradients of $g_t(\cdot)$ can be formulated as
 \begin{equation}\label{eq:gradienteffsurr}
     \nabla g_t(\y) = \left\{
\begin{array}{ll}
\nabla f_t(\x_t), & \text{if } \langle \nabla f_t(\x_t),\v_t \rangle \geq 0, \vspace{1mm}\\
\nabla f_t(\x_t) - \langle \nabla f_t(\x_t),\v_t \rangle \cdot \frac{\y- \Pi_\X [\y]}{\Vert \y- \Pi_\X [\y]\Vert},  & \text{if } \langle \nabla f_t(\x_t),\v_t \rangle < 0. 
\end{array} \right.
 \end{equation}
(i) When $\langle \nabla f_t(\x_t),\v_t \rangle \geq 0$. We have $g_t(\y)=\langle\nabla f_t(\x_t),\y \rangle$ and $\nabla g_t(\y) = \nabla f_t(\x_t)$. Thus,  
\begin{equation}\label{eqn:relationship1}
    \langle \nabla f_t(\x_t),\x_t-\x \rangle = \langle \nabla g_t(\y_t),\y_t-\x\rangle  - \langle\nabla f_t(\x_t),\y_t-\x_t \rangle. 
\end{equation}
By the definition of $\v_t=(\y_t-\x_t)/\Vert  
\y_t-\x_t \Vert$, we have $\langle \nabla f_t(\x_t),\x_t\rangle\leq \langle \nabla f_t(\x_t),\y_t\rangle$ and thus 
\begin{equation}\label{eqn:relationship3}
    \langle \nabla g_t(\y_t),\x_t\rangle\leq \langle \nabla g_t(\y_t),\y_t\rangle
\end{equation} 
(ii) When $\langle \nabla f_t(\x_t),\v_t \rangle < 0$. According to Lemma~\ref{thm:effsurr}, we obtain 
\begin{equation}\label{eqn:relationship2}
    \langle \nabla f_t(\x_t),\x_t-\x \rangle \leq \langle \nabla g_t(\y_t),\y_t-\x\rangle.  
\end{equation}
Moreover, we derive the following equation
\begin{equation}\label{eqn:relationship4}
    \begin{aligned}
        &\langle \nabla g_t(\y_t), \y_t - \x_t\rangle 
        = \langle \nabla f_t(\x_t), \y_t - \x_t\rangle - \langle \nabla f_t(\x_t),\v_t \rangle \cdot \langle \v_t, \y_t - \x_t \rangle \\
        = {} & \langle \nabla f_t(\x_t), \y_t - \x_t\rangle - \langle \nabla f_t(\x_t), \y_t - \x_t\rangle \cdot \frac{1}{\Vert \y_t - \x_t \Vert} \left\langle \frac{\y_t - \x_t}{\Vert \y_t - \x_t \Vert}, \y_t - \x_t\right\rangle 
        =0. 
    \end{aligned}
\end{equation}
Finally, combining \eqref{eqn:relationship1} and \eqref{eqn:relationship2} obtains \eqref{eqn:effsur:leqimprove}, further combining \eqref{eqn:relationship3} and \eqref{eqn:relationship4} 
yields \eqref{eqn:con:xtyt}. 

\subsection{Proof of Lemma~\ref{lem:meta-regret:effsurr}}
By the regret guarantee of Adapt-ML-Prod \citep[Corollary~4]{pmlr-v35-gaillard14}, we have 
\begin{equation*}
    \sum_{t=1}^T \left( \ell_t - \ell_t^i \right) \leq 2\Gamma + \frac{\Gamma}{\sqrt{\ln |\A|}}\sqrt{1+ \sum_{t=1}^T \left( \ell_t - \ell_t^i \right)^2 }
\end{equation*} 
for all expert $E^i\in\A$, where $\Gamma = 3\ln |\A|+\ln (1+\frac{|\A|}{2e}(1+\ln(T+1)))=O(\log\log T)$. By the definition of $\ell_t$ and $\ell_t^i$, we have 
\begin{equation}\label{eqn:second-oder:effsurr}
    \begin{aligned}
        &\sum_{t=1}^T \langle \nabla g_t(\y_t),\y_t-\y_t^i \rangle \leq 8\Gamma GD +\frac{\Gamma}{\sqrt{\ln |\A|}}  \sqrt{16G^2D^2 + \sum_{t=1}^T \langle \nabla g_t(\y_t),\y_t-\y_t^i \rangle^2 } \\
        \leq {} & 4\Gamma GD \left( 2+ \frac{1}{\sqrt{\ln |\A|}} \right) + \frac{\Gamma^2G^2}{2\gamma \ln |\A|} + \frac{\gamma}{2G^2} \sum_{t=1}^T \langle \nabla g_t(\y_t),\y_t-\y_t^i \rangle^2, 
    \end{aligned}
\end{equation}
for any $\gamma>0$, where the last step uses AM-GM inequality. 

Next, we handle the term $\langle \nabla g_t(\y_t), \y_t-\y_t^i\rangle^2$. We will consider two cases separately. 

(i) When $\langle \nabla f_t(\x_t),\v_t \rangle \geq 0$ , we have 
\begin{equation}\label{eqn:Lemma:vt1}
    \langle \nabla f_t(\x_t), \x_t-\y_t^i\rangle\leq \langle \nabla f_t(\x_t), \y_t-\y_t^i\rangle \leq \Vert \nabla f_t(\x_t)\Vert \Vert \y_t-\y_t^i\Vert  \leq 2GD.
\end{equation}
As the function $q(x)=x-\frac{\gamma}{2G^2}x^2$ is strictly increasing when $x\in (-\infty, \frac{G^2}{\gamma}]$, \eqref{eqn:Lemma:vt1} implies that
\begin{equation*}
    \langle \nabla f_t(\x_t), \x_t-\y_t^i\rangle
    - \frac{\gamma}{2G^2}\langle \nabla f_t(\x_t), \x_t-\y_t^i\rangle^2
    \leq \langle \nabla f_t(\x_t), \y_t-\y^i_t\rangle - \frac{\gamma}{2G^2}\langle \nabla f_t(\x_t), \y_t-\y^i_t\rangle^2. 
\end{equation*}
for any $\gamma\in (0,\frac{G}{2D}]$. By rearranging terms, we obtain
\begin{equation}\label{eqn:tl:xtyt1}
\begin{aligned}
    &\frac{\gamma}{2G^2}\langle \nabla g_t(\y_t), \y_t-\y_t^i\rangle^2 \overset{\eqref{eq:gradienteffsurr}}{=} \frac{\gamma}{2G^2}\langle \nabla f_t(\x_t), \y_t-\y_t^i\rangle^2 \\
    \leq {} & \langle \nabla f_t(\x_t), \y_t-\x_t\rangle + \frac{\gamma}{2G^2}\langle \nabla f_t(\x_t), \x_t-\y_t^i\rangle^2 \\ 
    \overset{\eqref{eq:gradienteffsurr}}{=} {} & \langle \nabla f_t(\x_t), \y_t-\x_t\rangle + \frac{\gamma}{2G^2}\langle \nabla g_t(\y_t), \x_t-\y_t^i\rangle^2. 
\end{aligned}
\end{equation}

(ii) When $\langle \nabla f_t(\x_t),\v_t \rangle < 0$, \eqref{eqn:relationship4} implies $\langle \nabla g_t(\y_t), \x_t - \y^i_t\rangle = \langle \nabla g_t(\y_t), \y_t - \y^i_t\rangle$. Thus, 
\begin{equation}\label{eqn:tl:xtyt2}
    \frac{\gamma}{2G^2}\langle \nabla g_t(\y_t), \y_t-\y_t^i\rangle^2 = \frac{\gamma}{2G^2}\langle \nabla g_t(\y_t), \x_t-\y_t^i\rangle^2. 
\end{equation}
Combining \eqref{eqn:tl:xtyt1} and \eqref{eqn:tl:xtyt2}, we have
\begin{equation}\label{eqn:tl:final}
    \frac{\gamma}{2G^2}\langle \nabla g_t(\y_t), \y_t-\y_t^i\rangle^2 
    \leq \mathds{1}_{\{\langle \nabla f_t(\x_t),\v_t \rangle \geq 0\}} \langle \nabla f_t(\x_t), \y_t-\x_t\rangle + \frac{\gamma}{2G^2}\langle \nabla g_t(\y_t), \x_t-\y_t^i\rangle^2
\end{equation}
for any $\gamma\in (0,\frac{G}{2D}]$. Substituting \eqref{eqn:tl:final} into \eqref{eqn:second-oder:effsurr}, we finish the proof. 

\subsection{Proof of Theorem~\ref{thm:effusc}}
We present the exact bounds of the theoretical guarantee provided in Theorem~\ref{thm:effusc}. When functions are general convex, we have
\begin{equation*}
\begin{aligned}
     \sum_{t=1}^T f_t(\x_t) - \sum_{t=1}^T f_t(\x)&\leq 4\Gamma GD \left( 2+ \frac{1}{\sqrt{\ln |\A|}} \right) + \left(\frac{2\Gamma GD}{\sqrt{\ln |\A|}}+2D^2+G^2 \right) \sqrt{ T } -\frac{G^2}{2} \\
     &= O(\sqrt{T})
\end{aligned}
\end{equation*}
where $|\A| = 1+2\lceil \log_2 T\rceil$ and 
\begin{equation}\label{eq:Gamma}
    \Gamma = 3\ln |\A|+\ln \left(1+\frac{|\A|}{2e}(1+\ln(T+1))\right)=O(\log\log T). 
\end{equation}
When functions are $\alpha$-exp-concave, we have
\begin{equation*}
\begin{aligned}
    \sum_{t=1}^T f_t(\x_t) - \sum_{t=1}^T f_t(\x) 
        &\leq  4\Gamma GD \left( 2+ \frac{1}{\sqrt{\ln |\A|}} \right) +\frac{\Gamma^2}{\beta\ln |\A|} + 5 \left(\frac{8}{\beta} +2\sqrt{2}GD\right) d\log T  \\
        &= O\left(\frac{d}{\alpha} \log T\right). 
\end{aligned}
\end{equation*}
When functions are $\lambda$-strongly convex, we have
\begin{equation*}
    \begin{aligned}
        \sum_{t=1}^T f_t(\x_t) - \sum_{t=1}^T f_t(\x)  &\leq 4\Gamma GD \left( 2+\frac{1}{\sqrt{\ln |\A|}} \right) + \frac{\Gamma^2 G^2}{\min\{\frac{G}{D},\lambda\} \ln |\A|} + \frac{(G+D)^2}{\lambda}\log T \\
        &= O\left(\frac{1}{\lambda}\log T\right). 
    \end{aligned}
\end{equation*}

\subsubsection{Analysis for General Convex Functions}
We introduce the following decomposition for general convex functions, 
\begin{equation}\label{eqn:ana:con:decom}
    \begin{aligned}
        \sum_{t=1}^T f_t(\x_t) - \sum_{t=1}^T f_t(\x) &\leq \sum_{t=1}^T \langle \nabla f_t(\x_t),\x_t-\x\rangle 
        \overset{\eqref{eqn:effsurr:leq}}{\leq} \sum_{t=1}^T \langle \nabla g_t(\y_t),\y_t-\x\rangle \\
        & = \sum_{t=1}^T \langle \nabla g_t(\y_t),\y_t-\y_t^i\rangle + \sum_{t=1}^T \langle \nabla g_t(\y_t),\y_t^i-\x\rangle \\
        & \overset{\eqref{eq:effUSC:surr:con}}{=} \underbrace{\sum_{t=1}^T \langle \nabla g_t(\y_t),\y_t-\y_t^i\rangle}_{\meta} + \underbrace{\sum_{t=1}^T \left( \ell_t^{\con} (\y_t^i) -\ell_t^{\con} (\x)\right)}_{\expert}. 
    \end{aligned}
\end{equation}
First, we start with the expert-regret. Since we are employing OGD to minimize $\ell_t^{\con} (\cdot)$, using standard OGD analysis~\citep[Theorem~1]{zinkevich-2003-online} can obtain the following upper bound
\begin{equation}
\label{eqn:ana:con:expert-regret}
    \sum_{t=1}^T  \ell_t^{\conn} (\y_t^i) -\sum_{t=1}^T\ell_t^{\conn} (\x) \leq (2D^2+G^2)\sqrt{T} - \frac{G^2}{2},
\end{equation}
for any expert $\y_t^i\in\Y$ and any $\x\in\X$. 

Next, we move to bound the meta-regret. According to \eqref{eqn:second-oder:effsurr}, we have
\begin{equation}\label{eqn:ana:con:meta-regret}
    \begin{aligned}
        \sum_{t=1}^T \langle \nabla g_t(\y_t),\y_t-\y_t^i \rangle &\leq 8\Gamma GD +\frac{\Gamma}{\sqrt{\ln |\A|}}  \sqrt{16G^2D^2 + \sum_{t=1}^T \langle \nabla g_t(\y_t),\y_t-\y_t^i \rangle^2 } \\
        &\leq 4\Gamma GD \left( 2+ \frac{1}{\sqrt{\ln |\A|}} \right) +\frac{\Gamma}{\sqrt{\ln |\A|}}  \sqrt{ \sum_{t=1}^T \langle \nabla g_t(\y_t),\y_t-\y_t^i \rangle^2 } \\
        &\leq  4\Gamma GD \left( 2+ \frac{1}{\sqrt{\ln |\A|}} \right) +\frac{\Gamma}{\sqrt{\ln |\A|}}  \sqrt{ \sum_{t=1}^T \Vert \nabla g_t(\y_t)\Vert^2 \Vert\y_t-\y_t^i \Vert^2 } \\
        &\leq 4\Gamma GD \left( 2+ \frac{1}{\sqrt{\ln |\A|}} \right) +\frac{2\Gamma GD}{\sqrt{\ln |\A|}}  \sqrt{ T },
    \end{aligned}
\end{equation}
for all expert $E^i\in\A$, where $\Gamma$ is defined in \eqref{eq:Gamma} and the last set is due to
\begin{equation}\label{eqn:effsurr:G}
    \Vert \nabla g_t(\y_t)\Vert\leq \Vert 
\nabla f_t(\x_t)\Vert\leq G. 
\end{equation}

Finally, substituting \eqref{eqn:ana:con:expert-regret} and \eqref{eqn:ana:con:meta-regret} into \eqref{eqn:ana:con:decom}, we have
\begin{equation*}
    \sum_{t=1}^T f_t(\x_t) - \sum_{t=1}^T f_t(\x)\leq 4\Gamma GD \left( 2+ \frac{1}{\sqrt{\ln |\A|}} \right) + \left(\frac{2\Gamma GD}{\sqrt{\ln |\A|}}+2D^2+G^2 \right) \sqrt{ T } -\frac{G^2}{2}. 
\end{equation*}

\subsubsection{Analysis for Exp-concave Functions}
For $\alpha$-exp-concave functions, there exits $\hat{\alpha}^*\in \P_{\expc}$ that $\hat{\alpha}^*\leq \alpha \leq 2\hat{\alpha}^*$, where $\hat{\alpha}^*$ is the modulus of the $i$-th expert $E^i$. This inequality also indicates
\begin{equation}\label{eqn:ana:exp:alpha}
    \hat{\beta}^*\leq \beta \leq 2\hat{\beta}^*, \quad \hat{\beta}^* = \frac{1}{2} \min \{\frac{1}{4GD},\hat{\alpha}^*\}. 
\end{equation}
Since $x-\frac{\hat{\beta}^*}{2}x^2$ is strictly increasing where $\hat{\beta}^*=\frac{1}{2}\min \{\frac{1}{4GD},\hat{\alpha}^*\}$ when $x\in (-\infty,2GD]$, \eqref{eqn:effsurr:leq} implies that
\begin{equation}\label{eqn:ana:exp:ineq}
    \langle \nabla f_t(\x_t),\x_t-\x\rangle - \frac{\hat{\beta}^*}{2} \langle \nabla f_t(\x_t),\x_t-\x\rangle^2 \leq \langle \nabla g_t(\y_t),\y_t-\x\rangle - \frac{\hat{\beta}^*}{2} \langle \nabla g_t(\y_t),\y_t-\x\rangle^2. 
\end{equation}
Then, we introduce the following decomposition for $\alpha$-exp-concave functions, 
\begin{equation}\label{eqn:ana:exp:decom}
    \begin{aligned}
        &\sum_{t=1}^T f_t(\x_t) - \sum_{t=1}^T f_t(\x) \leq \sum_{t=1}^T \langle \nabla f_t(\x_t),\x_t-\x\rangle - \frac{\beta}{2} \sum_{t=1}^T \langle \nabla f_t(\x_t),\x_t-\x\rangle^2 \\
        \overset{\eqref{eqn:ana:exp:alpha}}{\leq} {} & \sum_{t=1}^T \langle \nabla f_t(\x_t),\x_t-\x\rangle - \frac{\hat{\beta}^*}{2} \sum_{t=1}^T \langle \nabla f_t(\x_t),\x_t-\x\rangle^2 \\
        \overset{\eqref{eqn:ana:exp:ineq}}{\leq} {} & \sum_{t=1}^T \langle \nabla g_t(\y_t),\y_t-\x\rangle - \frac{\hat{\beta}^*}{2} \sum_{t=1}^T \langle \nabla g_t(\y_t),\y_t-\x\rangle^2 \\
         = {} & \sum_{t=1}^T \langle \nabla g_t(\y_t),\y_t-\y_t^i\rangle + \sum_{t=1}^T \langle \nabla g_t(\y_t),\y_t^i-\x\rangle - \frac{\hat{\beta}^*}{2} \sum_{t=1}^T \langle \nabla g_t(\y_t),\y_t-\x\rangle^2 \\
         \overset{\eqref{eq:effUSC:surr:exp}}{=} {} & \underbrace{\sum_{t=1}^T \langle \nabla g_t(\y_t),\y_t-\y_t^i\rangle}_{\meta} + \underbrace{\sum_{t=1}^T \left( \ell_{t,\hat{\alpha}^*}^{\expc} (\y_t^i) -\ell_{t,\hat{\alpha}^*}^{\expc} (\x)\right)}_{\expert} - \frac{\hat{\beta}^*}{2} \sum_{t=1}^T \langle \nabla g_t(\y_t),\y_t-\y_t^i\rangle^2. 
    \end{aligned}
\end{equation}
For the expert-regret, we can use the analysis of ONS \citep[Theorem~2]{ML:Hazan:2007} to obtain
\begin{equation}\label{eqn:ana:exp:expert-regret}
    \sum_{t=1}^T  \ell_{t,\hat{\alpha}^*}^{\expn} (\y_t^i) -\sum_{t=1}^T\ell_{t,\hat{\alpha}^*}^{\expn} (\x) \leq 5 \left(\frac{4}{\hat{\beta}^*} +2\sqrt{2}GD\right) d\log T 
\end{equation}
for any expert $\y_t^i\in \Y$ and any $\x\in\X$, where $\hat{\beta}^*$ is defined in \eqref{eqn:ana:exp:alpha}. 
Next, we move to bound the meta-regret. According to \eqref{eqn:second-oder:effsurr}, we have
\begin{equation}\label{eqn:ana:exp:meta-regret}
    \begin{aligned}
        \sum_{t=1}^T \langle \nabla g_t(\y_t),\y_t-\y_t^i \rangle &\leq 8\Gamma GD +\frac{\Gamma}{\sqrt{\ln |\A|}}  \sqrt{16G^2D^2 + \sum_{t=1}^T \langle \nabla g_t(\y_t),\y_t-\y_t^i \rangle^2 } \\
        &\leq 4\Gamma GD \left( 2+ \frac{1}{\sqrt{\ln |\A|}} \right) +\frac{\Gamma}{\sqrt{\ln |\A|}}  \sqrt{ \sum_{t=1}^T \langle \nabla g_t(\y_t),\y_t-\y_t^i \rangle^2 } \\
        &\leq  4\Gamma GD \left( 2+ \frac{1}{\sqrt{\ln |\A|}} \right) +\frac{\Gamma^2}{2\hat{\beta}^*\ln |\A|} + \frac{\hat{\beta}^*}{2} \langle \nabla g_t(\y_t),\y_t-\y_t^i \rangle^2 
    \end{aligned}
\end{equation}
for all expert $E^i\in\A$, where $\Gamma$ is defined in \eqref{eq:Gamma} and the last step is due to $\sqrt{ab}\leq \frac{a}{2}+\frac{b}{2}$. Substituting  \eqref{eqn:ana:exp:expert-regret} and \eqref{eqn:ana:exp:meta-regret} into \eqref{eqn:ana:exp:decom}, we have
\begin{equation*}
        \sum_{t=1}^T f_t(\x_t) - \sum_{t=1}^T f_t(\x) 
        \leq  4\Gamma GD \left( 2+ \frac{1}{\sqrt{\ln |\A|}} \right) +\frac{\Gamma^2}{2\hat{\beta}^*\ln |\A|} + 5 \left(\frac{4}{\hat{\beta}^*} +2\sqrt{2}GD\right) d\log T .
\end{equation*}
Finally, we use \eqref{eqn:ana:exp:alpha} to simplify the above bound. 

\subsubsection{Analysis for Strongly Convex Functions}
For $\lambda$-strongly convex functions, there exits $\hat{\lambda}^*\in \P_{\strc}$ that $\hat{\lambda}^*\leq \lambda \leq 2\hat{\lambda}^*$, where $\hat{\lambda}^*$ is the modulus of the $i$-th expert $E^i$. Then, we introduce the following decomposition for $\lambda$-strongly convex functions
\begin{equation}\label{eqn:ana:str:decom}
    \begin{aligned}
        &\sum_{t=1}^T f_t(\x_t) - \sum_{t=1}^T f_t(\x) \leq \sum_{t=1}^T \langle \nabla f_t(\x_t),\x_t-\x\rangle - \frac{\lambda}{2} \sum_{t=1}^T \Vert\x_t-\x\Vert^2 \\
        \leq&  \sum_{t=1}^T \langle \nabla f_t(\x_t),\x_t-\x\rangle - \frac{\hat{\lambda}^*}{2} \sum_{t=1}^T \Vert\x_t-\x\Vert^2 \\
        \overset{\eqref{eqn:effsur:leqimprove}}{\leq}& \sum_{t=1}^T \langle \nabla g_t(\y_t),\y_t-\x\rangle -\Delta_T - \frac{\hat{\lambda}^*}{2} \sum_{t=1}^T \Vert\x_t-\x\Vert^2 \\
        \overset{\eqref{eq:effUSC:surr:str}}{=}& \underbrace{\sum_{t=1}^T \langle \nabla g_t(\y_t),\y_t-\y_t^i\rangle}_{\meta} + \underbrace{\sum_{t=1}^T \left( \ell_{t,\hat{\lambda}^*}^{\strc} (\y_t^i) -\ell_{t,\hat{\lambda}^*}^{\strc} (\x)\right)}_{\expert} - \frac{\hat{\lambda}^*}{2} \sum_{t=1}^T \Vert\y^i_t-\x_t\Vert^2-\Delta_T
    \end{aligned}
\end{equation}
where $\Delta_T = \sum_{t=1}^T \mathds{1}_{\{ \langle\nabla f_t(\x_t),\v_t\rangle\geq 0 \}} \cdot \langle\nabla f_t(\x_t),\y_t-\x_t \rangle$. To bound the meta-regret, we combine Lemma~\ref{lem:meta-regret:effsurr} with \eqref{eqn:ana:str:decom} to attain 
\begin{equation}\label{eqn:ana:str:decom2}
    \begin{aligned}
        &\sum_{t=1}^T f_t(\x_t) - \sum_{t=1}^T f_t(\x)  \\
        \leq {} &  4\Gamma GD \left( 2+\frac{1}{\sqrt{\ln |\A|}} \right) + \frac{\Gamma^2 G^2}{2\gamma \ln |\A|} + \frac{\gamma}{2G^2} \sum_{t=1}^T \langle \nabla g_t(\y_t),\x_t-\y_t^i \rangle^2 \\
        &+ \text{ER}(T) - \frac{\hat{\lambda}^*}{2} \sum_{t=1}^T \Vert\y^i_t-\x_t\Vert^2 \\
        \overset{\eqref{eqn:effsurr:G}}{\leq}&  4\Gamma GD \left( 2+\frac{1}{\sqrt{\ln |\A|}} \right) + \frac{\Gamma^2 G^2}{2\gamma \ln |\A|} + \left(\frac{\gamma}{2}-\frac{\hat{\lambda}^*}{2} \right)\sum_{t=1}^T \Vert\x_t-\y_t^i \Vert^2 + \text{ER}(T) \\
        \leq {} & 4\Gamma GD \left( 2+\frac{1}{\sqrt{\ln |\A|}} \right) + \frac{\Gamma^2 G^2}{2\gamma \ln |\A|}  + \text{ER}(T)
    \end{aligned}
\end{equation}
where $\text{ER}(T) = \sum_{t=1}^T ( \ell_{t,\hat{\lambda}^*}^{\strc} (\y_t^i) -\ell_{t,\hat{\lambda}^*}^{\strc} (\x))$ and the last step is because we set $\gamma=\min\{\frac{G}{2D},\hat{\lambda}^*\}$. 
Next, we move to bound the expert-regret by utilizing standard analysis of OGD \citep[Lemma~1]{Shalev-ShwartzSSC11}
\begin{equation}\label{eqn:ana:str:expert-regret}
    \textnormal{ER}(T)=\sum_{t=1}^T  \ell_{t,\hat{\lambda}^*}^{\strn} (\y_t^i) -\sum_{t=1}^T\ell_{t,\hat{\lambda}^*}^{\strn} (\x) \leq \frac{(G+D)^2}{2\hat{\lambda}^*}\log T.  
\end{equation}
for any expert $\y_t^i\in \Y$ and any $\x\in\X$. 
Substituting \eqref{eqn:ana:str:expert-regret} into \eqref{eqn:ana:str:decom2}, we have
\begin{equation*}
    \sum_{t=1}^T f_t(\x_t) - \sum_{t=1}^T f_t(\x)  \leq 4\Gamma GD \left( 2+\frac{1}{\sqrt{\ln |\A|}} \right) + \frac{\Gamma^2 G^2}{2\gamma \ln |\A|} + \frac{(G+D)^2}{2\hat{\lambda}^*}\log T. 
\end{equation*}
Finally, we use $\hat{\lambda}^*\leq \lambda \leq 2\hat{\lambda}^*$ to simplify the above bound. 

\subsection{Proof of Theorem~\ref{thm:effusc:smooth}}
The analysis is similar to Theorem~\ref{thm:effusc}. Also, we present the exact bounds of the theoretical guarantee provided in Theorem~\ref{thm:effusc:smooth}. When functions are general convex, we have
\begin{equation*}
\begin{aligned}
    &\sum_{t=1}^T f_t(\x_t) - \sum_{t=1}^T f_t(\x) \\
   \leq {} & 4\Gamma GD \left( 2+ \frac{1}{\sqrt{\ln |\A|}} \right) + \sqrt{2D^2\delta} + 4H \left( \frac{2\Gamma D}{\sqrt{\ln |\A|}} + \sqrt{2} (D+2G) \right)^2  \\
    &+ 2\sqrt{H}\left( \frac{2\Gamma D}{\sqrt{\ln |\A|}} + \sqrt{2} (D+2G) \right)\sqrt{ L_T + 4\Gamma GD \left( 2+ \frac{1}{\sqrt{\ln |\A|}} \right) + \sqrt{2D^2\delta} } \\
    = {} & O(\sqrt{L_T}). 
\end{aligned}
\end{equation*}
where $|\A| = 1+2\lceil \log_2 T\rceil$, $\Gamma$ is defined in \eqref{eq:Gamma}, and $L_T=\min_{\x\in\X} \sum_{t=1}^T f_t(\x)$.  
When functions are $\alpha$-exp-concave, we have
\begin{equation*}
\begin{aligned}
    &\sum_{t=1}^T f_t(\x_t) - \sum_{t=1}^T f_t(\x)  \\
         \leq {} & 4\Gamma GD \left( 2+ \frac{1}{\sqrt{\ln |\A|}} \right) +\frac{\Gamma^2}{2\beta\ln |\A|} + \frac{2d}{\beta} \log \left( \frac{\beta^2D^2H}{d} \sum_{t=1}^{T} f_t(\x_t) +1\right)+\frac{2}{\beta} \\
         \leq {} & \hat{\Gamma}
         + \frac{2d}{\beta} \log \left( \frac{2\beta^2 D^2H}{d} \sum_{t=1}^{T} f_t(\x) + \frac{2\beta^2D^2H}{d} \hat{\Gamma} + 2D^2H\log (2D^2H)+2\right) \\
         = {} & O\left(\frac{d}{\alpha}\log L_T\right)
\end{aligned}
\end{equation*}
where $\hat{\Gamma} = 4\Gamma GD \left( 2+ \frac{1}{\sqrt{\ln |\A|}} \right) +\frac{\Gamma^2}{2\beta\ln |\A|} +\frac{2}{\beta}$. 
When functions are $\lambda$-strongly convex, we have
\begin{equation*}
    \begin{aligned}
        &\sum_{t=1}^T f_t(\x_t) - \sum_{t=1}^T f_t(\x) \\
        \leq {} &  \tilde{\Gamma} +\frac{(G+2D)^2}{2\lambda} \log \left(\frac{8H\lambda}{(G+2D)^2} \sum_{t=1}^{T} f_t(\x) +\frac{8H\lambda}{(G+2D)^2}\tilde{\Gamma}+2H\log(2H)+2\right) \\
        = {} & O\left(\frac{1}{\lambda}\log L_T\right)
    \end{aligned}
\end{equation*}
where $\tilde{\Gamma} = 4\Gamma GD \left( 2+\frac{1}{\sqrt{\ln |\A|}} \right) + \frac{\Gamma^2 G^2}{2\gamma \ln |\A|} + 1$. 

\subsubsection{Analysis for General Convex Functions}
We start with the meta-expert regret decomposition as presented in \eqref{eqn:ana:con:decom}, 
\begin{equation}\label{eqn:ana:con:decom:smooth}
        \sum_{t=1}^T f_t(\x_t) - \sum_{t=1}^T f_t(\x) 
         \leq \underbrace{\sum_{t=1}^T \langle \nabla g_t(\y_t),\y_t-\y_t^i\rangle}_{\meta} + \underbrace{\sum_{t=1}^T \left( \ell_t^{\con} (\y_t^i) -\ell_t^{\con} (\x)\right)}_{\expert}. 
\end{equation}
For the meta-regret, we reuse \eqref{eqn:ana:con:meta-regret} to obtain
\begin{equation}\label{eqn:ana:con:meta-regret:smooth}
    \begin{aligned}
        \sum_{t=1}^T \langle \nabla g_t(\y_t),\y_t-\y_t^i \rangle 
        &\leq  4\Gamma GD \left( 2+ \frac{1}{\sqrt{\ln |\A|}} \right) +\frac{\Gamma}{\sqrt{\ln |\A|}}  \sqrt{ \sum_{t=1}^T \Vert \nabla g_t(\y_t)\Vert^2 \Vert\y_t-\y_t^i \Vert^2 } \\
        &\leq 4\Gamma GD \left( 2+ \frac{1}{\sqrt{\ln |\A|}} \right) +\frac{2\Gamma D}{\sqrt{\ln |\A|}}  \sqrt{ \sum_{t=1}^T \Vert \nabla g_t(\y_t)\Vert^2 },
    \end{aligned}
\end{equation}
for all expert $E^i\in\A$, where $\Gamma$ is defined in \eqref{eq:Gamma}. For the expert-regret, we can use the analysis of SOGD \citep[Theorem~2]{Adaptive:Regret:Smooth:ICML} to obtain
\begin{equation*}
        \sum_{t=1}^T \ell^{\conn}_t (\y_t^i) - \sum_{t=1}^T \ell^{\conn}_t (\x) \leq \sqrt{2D^2} \sqrt{\delta + \left(1+\frac{2G}{D}\right)^2 \sum_{t=1}^T \Vert \nabla g_t(\y_t)\Vert^2 }. 
\end{equation*}
for any expert $\y_t^i\in\Y$ and any $\x\in\X$. 
From the above formulation, we have
\begin{equation}\label{eqn:ana:con:expert-regret:smooth}
        \sum_{t=1}^T \ell^{\conn}_t (\y_t^i) - \sum_{t=1}^T \ell^{\conn}_t (\x) \leq \sqrt{2D^2\delta} + \sqrt{2\left(D+2G\right)^2 \sum_{t=1}^T \Vert \nabla g_t(\y_t)\Vert^2 }. 
    \end{equation}
Substituting \eqref{eqn:ana:con:meta-regret:smooth} and \eqref{eqn:ana:con:expert-regret:smooth} into \eqref{eqn:ana:con:decom:smooth}, we have
\begin{equation*}
\begin{aligned}
    &\sum_{t=1}^T f_t(\x_t) - \sum_{t=1}^T f_t(\x) \\
    \overset{\eqref{eqn:effsurr:G}}{\leq} {} & 4\Gamma GD \left( 2+ \frac{1}{\sqrt{\ln |\A|}} \right) + \sqrt{2D^2\delta} + \left( \frac{2\Gamma D}{\sqrt{\ln |\A|}} + \sqrt{2} (D+2G) \right)\sqrt{ \sum_{t=1}^T \Vert \nabla f_t(\x_t)\Vert^2 }.
\end{aligned}
\end{equation*}
Next, we introduce the self-bounding property of smooth functions. 
\begin{lem}[{Lemma~3.1 of~\citet{NIPS2010_Smooth}}]
\label{lem:H-smooth}
    For an $H$-smooth and nonnegative function, we have $\Vert \nabla f(\x)\Vert\leq \sqrt{4Hf(\x)}$. 
\end{lem}
Thus, when functions are smooth, we have 
\begin{equation*}
\begin{aligned}
    &\sum_{t=1}^T f_t(\x_t) - \sum_{t=1}^T f_t(\x) \\
    \overset{\eqref{eqn:effsurr:G}}{\leq} & 4\Gamma GD \left( 2+ \frac{1}{\sqrt{\ln |\A|}} \right) + \sqrt{2D^2\delta} + \left( \frac{2\Gamma D}{\sqrt{\ln |\A|}} + \sqrt{2} (D+2G) \right)\sqrt{ 4H\sum_{t=1}^T f_t(\x_t) }.
\end{aligned}
\end{equation*}
To simplify the above inequality, we use the following lemma. 
\begin{lem}[{Lemma~19 of \citet{Shai:thesis}}]
\label{lem:xbc}
    Let $x,b,c\in \R^+$. Then, we have $x-c\leq b\sqrt{x} \Rightarrow x-c\leq b^2+b\sqrt{c}$. 
\end{lem}
By utilizing Lemma~\ref{lem:xbc}, we finish the proof. 

\subsubsection{Analysis for Exp-concave Functions}
The analysis is also similar to Theorem~\ref{thm:effusc}. We start with \eqref{eqn:ana:exp:decom}
\begin{equation}\label{eqn:ana:exp:decom:smooth}
\begin{aligned}
    &\sum_{t=1}^T f_t(\x_t) - \sum_{t=1}^T f_t(\x) \\
         \leq&  \underbrace{\sum_{t=1}^T \langle \nabla g_t(\y_t),\y_t-\y_t^i\rangle}_{\meta} + \underbrace{\sum_{t=1}^T \left( \ell_{t,\hat{\alpha}^*}^{\expc} (\y_t^i) -\ell_{t,\hat{\alpha}^*}^{\expc} (\x)\right)}_{\expert} - \frac{\hat{\beta}^*}{2} \sum_{t=1}^T \langle \nabla g_t(\y_t),\y_t-\y_t^i\rangle^2. 
\end{aligned}
\end{equation}
For the meta-regret, we also use \eqref{eqn:ana:exp:meta-regret} to bound. For the expert-regret, we can use the analysis of ONS under the smoothness condition \citep[Theorem~1]{Beyond:Logarithmic} to get 
\begin{equation*}
        \sum_{t=1}^T \ell^{\expn}_{t,\hat{\alpha}} (\y_t^i) - \sum_{t=1}^T \ell^{\expn}_{t,\hat{\alpha}} (\x) \leq  \frac{2d}{\hat{\beta}^*} \log \left( \frac{\hat{\beta}^{*^2} D^2}{16d} \sum_{t=1}^{T} \Vert\nabla \ell^{\expn}_{t,\hat{\alpha}} (\y_t^i) \Vert^2 +1\right)+\frac{2}{\hat{\beta}^*}. 
\end{equation*}
for any expert $\y_t^i\in\Y$ and any $\x\in\X$. 
Next, we provide an upper bound for $\Vert\nabla \ell^{\expn}_{t,\hat{\alpha}} (\y_t^i) \Vert^2$ 
\begin{equation*}
    \begin{aligned}
        &\Vert\nabla \ell^{\expn}_{t,\hat{\alpha}} (\y_t^i) \Vert^2 \\
        = {} & \langle \nabla g_t(\y_t) + \hat{\beta}^* \nabla g_t(\y_t) \nabla g_t(\y_t)^\top(\y-\y_t), \nabla g_t(\y_t) + \hat{\beta}^* \nabla g_t(\y_t) \nabla g_t(\y_t)^\top(\y-\y_t)\rangle \\
        = {} & \Vert \nabla g_t(\y_t)\Vert^2 + 2 \hat{\beta}^* \langle \nabla g_t(\y_t),\y-\y_t\rangle \Vert \nabla g_t(\y_t)\Vert^2 + \hat{\beta}^{*^2} \Vert \nabla g_t(\y_t)\Vert^4 \Vert \y-\y_t\Vert^2 \\
        \leq {} & \left( 1+2\hat{\beta}^{*^2}GD \right)^2 \Vert \nabla g_t(\y_t)\Vert^2 \leq 4\Vert \nabla g_t(\y_t)\Vert^2. 
    \end{aligned}
\end{equation*}
Thus, we have
\begin{equation}\label{eqn:ana:exp:expert-regret:smooth}
        \sum_{t=1}^T \ell^{\expn}_{t,\hat{\alpha}} (\y_t^i) - \sum_{t=1}^T \ell^{\expn}_{t,\hat{\alpha}} (\x) \leq  \frac{2d}{\hat{\beta}^*} \log \left( \frac{\hat{\beta}^{*^2}D^2}{4d} \sum_{t=1}^{T} \Vert\nabla g_t(\y_t) \Vert^2 +1\right)+\frac{2}{\hat{\beta}^*}
\end{equation}
Substituting \eqref{eqn:ana:exp:meta-regret} and \eqref{eqn:ana:exp:expert-regret:smooth} into \eqref{eqn:ana:exp:decom:smooth}, we have
\begin{equation}
\begin{aligned}
    &\sum_{t=1}^T f_t(\x_t) - \sum_{t=1}^T f_t(\x) \\
         \leq {} &  4\Gamma GD \left( 2+ \frac{1}{\sqrt{\ln |\A|}} \right) +\frac{\Gamma^2}{2\hat{\beta}^*\ln |\A|} + \frac{2d}{\hat{\beta}^*} \log \left( \frac{\hat{\beta}^{*^2}D^2}{4d} \sum_{t=1}^{T} \Vert\nabla g_t(\y_t) \Vert^2 +1\right)+\frac{2}{\hat{\beta}^*} \\
         \overset{\eqref{eqn:effsurr:G}}{\leq} {} & 4\Gamma GD \left( 2+ \frac{1}{\sqrt{\ln |\A|}} \right) +\frac{\Gamma^2}{2\hat{\beta}^*\ln |\A|} + \frac{2d}{\hat{\beta}^*} \log \left( \frac{\hat{\beta}^{*^2}D^2H}{d} \sum_{t=1}^{T} f_t(\x_t) +1\right)+\frac{2}{\hat{\beta}^*}
\end{aligned}
\end{equation}
where the last step is due to Lemma~\ref{lem:H-smooth}. Finally, we use the following lemma to simplify the bound. 
\begin{lem}[{Corollary~5 of \citet{Beyond:Logarithmic}}]
\label{lem:ln}
    Let $a,b,c,d,x>0$ satisfy $x-d\leq a\ln (bx+c)$. Then, we have $x-d\leq a\ln (2(ab\ln \frac{2ab}{\mathrm{e}} +db+c ))$. 
\end{lem}

\subsubsection{Analysis for Strongly Convex Functions}
Recall that we construct the expert-loss for strongly convex functions as follows
\begin{equation*}
     \hat{\ell}^{\strc}_{t,\hat{\lambda}} (\y)= \langle \nabla g_t(\y_t),\y-\y_t\rangle + \frac{\hat{\lambda}^*}{2G^2} \Vert  \nabla g_t(\y_t) \Vert^2\Vert \y-\x_t\Vert^2.  
 \end{equation*}
Then, we introduce a new decomposition for $\lambda$-strongly convex functions
\begin{equation}\label{eqn:ana:str:decom:smooth}
    \begin{aligned}
        &\sum_{t=1}^T f_t(\x_t) - \sum_{t=1}^T f_t(\x) \leq \sum_{t=1}^T \langle \nabla f_t(\x_t),\x_t-\x\rangle - \frac{\lambda}{2} \sum_{t=1}^T \Vert\x_t-\x\Vert^2 \\
        \leq {} &  \sum_{t=1}^T \langle \nabla f_t(\x_t),\x_t-\x\rangle - \frac{\hat{\lambda}^*}{2} \sum_{t=1}^T \Vert\x_t-\x\Vert^2 \\
        \leq {} & \sum_{t=1}^T \langle \nabla f_t(\x_t),\x_t-\x\rangle - \frac{\hat{\lambda}^*}{2G^2} \sum_{t=1}^T\Vert  \nabla g_t(\y_t) \Vert^2 \Vert\x_t-\x\Vert^2 \\
        \overset{\eqref{eqn:effsur:leqimprove}}{\leq} {} & \sum_{t=1}^T \langle \nabla g_t(\y_t),\y_t-\x\rangle -\Delta_T - \frac{\hat{\lambda}^*}{2G^2} \sum_{t=1}^T\Vert  \nabla g_t(\y_t) \Vert^2 \Vert\x_t-\x\Vert^2 \\       \overset{\eqref{eq:effUSC:surr:str:small-loss}}{=} {} & \underbrace{\sum_{t=1}^T \langle \nabla g_t(\y_t),\y_t-\y_t^i\rangle}_{\meta} + \underbrace{\sum_{t=1}^T \left( \hat{\ell}_{t,\hat{\lambda}^*}^{\strc} (\y_t^i) -\hat{\ell}_{t,\hat{\lambda}^*}^{\strc} (\x)\right)}_{\expert} -  \frac{\hat{\lambda}^*}{2G^2} \sum_{t=1}^T\Vert  \nabla g_t(\y_t) \Vert^2 \Vert\x_t-\y_t^i\Vert^2-\Delta_T
    \end{aligned}
\end{equation}
where $\Delta_T = \sum_{t=1}^T \mathds{1}_{\{ \langle\nabla f_t(\x_t),\v_t\rangle\geq 0 \}} \cdot \langle\nabla f_t(\x_t),\y_t-\x_t \rangle$. To bound the meta-regret, we still incorporate with Lemma~\ref{lem:meta-regret:effsurr} to get 
\begin{equation*}
        \sum_{t=1}^T f_t(\x_t) - \sum_{t=1}^T f_t(\x) 
        \leq  4\Gamma GD \left( 2+\frac{1}{\sqrt{\ln |\A|}} \right) + \frac{\Gamma^2 G^2}{2\gamma \ln |\A|} + \sum_{t=1}^T \left(\hat{\ell}_{t,\hat{\lambda}^*}^{\strc} (\y_t^i) -\hat{\ell}_{t,\hat{\lambda}^*}^{\strc} (\x)\right). 
\end{equation*}
For the expert-regret, we derive a variant of theoretical guarantee of S$^2$OGD. 
\begin{lem}\label{lem:S2OGD}
    Under Assumptions~\ref{ass:D} and \ref{ass:G}, for any expert $\y_t^i\in\Y$ and any $\x\in\X$, we have
    \begin{equation*}
        \sum_{t=1}^T \hat{\ell}^{\strn}_{t,\hat{\lambda}^*} (\y_t^i) - \sum_{t=1}^T \hat{\ell}^{\strn}_{t,\hat{\lambda}^*} (\x) \leq   1+\frac{(G+2D)^2}{2\hat{\lambda}^*} \log \left(\frac{\hat{\lambda}^*}{(G+2D)^2} \sum_{t=1}^{T} \Vert \nabla g_t(\y_t)\Vert^2+1\right)
    \end{equation*} 
\end{lem}
Combining the above bounds, we have
\begin{equation*}
    \begin{aligned}
        &\sum_{t=1}^T f_t(\x_t) - \sum_{t=1}^T f_t(\x) \\
        \leq & 4\Gamma GD \left( 2+\frac{1}{\sqrt{\ln |\A|}} \right) + \frac{\Gamma^2 G^2}{2\gamma \ln |\A|} + 1+\frac{(G+2D)^2}{2\hat{\lambda}^*} \log \left(\frac{4H\hat{\lambda}^*}{(G+2D)^2} \sum_{t=1}^{T} f_t(\x)+1\right).  
    \end{aligned}
\end{equation*}
Finally, we simplify the above bound by utilizing Lemma~\ref{lem:ln}. 

\section{Conclusion and Future Work}
\label{sec:conclusion}
In this paper, we propose a projection-efficient universal algorithm that achieves minimax optimal regret for three types of convex functions with only $1$ projection per round. Furthermore, we enhance our algorithm to exploit the smoothness property and demonstrate that it attains small-loss regret for convex and smooth functions. 

One potentially unfavorable characteristic of our work is the requirements of domain and gradient boundedness. Given the recent developments in parameter-free online learning for \emph{unbounded domains and gradients}~\citep{NIPS2016_32072254,NIPS2016_550a141f,pmlr-v65-cutkosky17a,COLT'22:Corral,pmlr-v178-jacobsen22a}, in the future we will investigate whether our algorithms can further avoid prior knowledge of domain diameter $D$ and gradient norm upper bound $G$. 

Moreover, in addition to the small-loss bound, another important type of problem-dependent guarantee is the  \emph{gradient-variation regret bound}~\citep{NeurIPS'20:Sword,JMLR'24:Sword++}, which has been actively studied recently due to its profound relationship to games and stochastic optimization. Recently, \citet{NeurIPS'23:universal} achieve almost-optimal gradient-variation regret in universal online learning, but the algorithm maintains $O(\log^2 T)$ experts and conducts $O(\log^2 T)$ projections onto feasible domain $\X$ per round. Therefore, it remains challenging and important to develop a projection-efficient universal algorithm with optimal gradient-variation regret guarantees.

\bibliography{EffUni}

\newpage
\appendix

\section{Algorithms for Experts}\label{app:expert}
In this section, we provide the procedures of the expert-algorithms in our efficient algorithm. 

\subsection{Online Gradient Descent for Convex Functions} 
We use OGD~\citep{zinkevich-2003-online} to minimize $\ell_t^{\con}(\cdot)$ in \eqref{eq:effUSC:surr:con}. The procedure of the expert-algorithm for general convex functions is summarized in Algorithm~\ref{alg:ogd}. 
\begin{algorithm}[H]
   \caption{Expert $E^i$: OGD for Convex Functions}\label{alg:ogd}
\begin{algorithmic}[1]
   \STATE Let $\y^i_1$ be any point in $\Y$
   \FOR{$t=1$ \TO $T$} 
   \STATE Submit $\y^i_t$ to the meta-algorithm
   \STATE Update 
   \begin{equation*}
       \hat{\y}^i_{t+1} = \y^i_t - \frac{1}{\sqrt{t}} \nabla g_t(\y_t)
   \end{equation*}
   \STATE Conduct a projection onto $\Y$ 
   \begin{equation*}
    \y^i_{t+1}   = \left\{
\begin{array}{ll}
\hat{\y}^i_{t+1}, & \text{if } \Vert  \hat{\y}^i_{t+1}\Vert\leq D,\vspace{1mm}\\
\hat{\y}^i_{t+1}\cdot \frac{D}{\Vert  \hat{\y}^i_{t+1}\Vert},& \text{otherwise }.
\end{array} \right.
   \end{equation*}
   \ENDFOR
\end{algorithmic}
\end{algorithm}

\subsection{Online Gradient Descent for Strongly Convex Functions}
\begin{algorithm}[H]
   \caption{Expert $E^i$: OGD for Strongly Convex Functions}\label{alg:ogdstr}
\begin{algorithmic}[1]
   \STATE Let $\y^i_1$ be any point in $\Y$
   \FOR{$t=1$ \TO $T$} 
   \STATE Submit $\y^i_t$ to the meta-algorithm
   \STATE Update 
   \begin{equation*}
       \hat{\y}^i_{t+1} = \y^i_t - \frac{1}{\hat{\lambda}t} \nabla \ell^{\strc}_{t,\hat{\lambda}} (\y^i_t)
   \end{equation*}
   where
   \begin{equation*}
       \nabla \ell^{\strc}_{t,\hat{\lambda}} (\y^i_t) = \nabla g_t(\y_t) + \hat{\lambda} (\y^i_t-\x_t)
   \end{equation*}
   \STATE Conduct a projection onto $\Y$ 
   \begin{equation*}
    \y^i_{t+1}   = \left\{
\begin{array}{ll}
\hat{\y}^i_{t+1}, & \text{if } \Vert  \hat{\y}^i_{t+1}\Vert\leq D,\vspace{1mm}\\
\hat{\y}^i_{t+1}\cdot \frac{D}{\Vert  \hat{\y}^i_{t+1}\Vert},& \text{otherwise }.
\end{array} \right.
   \end{equation*}
   \ENDFOR
\end{algorithmic}
\end{algorithm}
We establish the following lemma for function $\ell_{t}^{\strc} (\cdot)$ in \eqref{eq:effsurr:str}. 
\begin{lem}\label{lem:surrstr}
     Under Assumptions~\ref{ass:D} and \ref{ass:G}, the loss function $\ell_{t}^{\strn} (\cdot)$ in \eqref{eq:effsurr:str} is $\lambda$-strongly convex, and $\Vert \nabla \ell_{t}^{\strn} (\y) \Vert^2  \leq (G+2D)^2$. 
 \end{lem}
Since $\ell^{\strc}_{t,\hat{\lambda}} (\cdot)$ in \eqref{eq:effUSC:surr:str} shares the same formulation as $\ell_{t}^{\strc} (\cdot)$, $\ell^{\strc}_{t,\hat{\lambda}} (\cdot)$ also benefits from the aforementioned properties, with the distinction being the substitution of $\lambda$ for $\hat{\lambda}$. Therefore, we use a variant of OGD~\citep{ICML_Pegasos} to minimize $\ell^{\strc}_{t,\hat{\lambda}} (\cdot)$. The procedure is summarized in Algorithm~\ref{alg:ogdstr}. 

\subsection{Online Newton Step for Exp-concave (and Smooth) Functions}
We establish the following lemma for functions $\ell^{\expc}_{t,\hat{\alpha}} (\cdot)$ in \eqref{eq:effUSC:surr:exp}. 
\begin{lem}\label{lem:surrexp}
     Under Assumptions~\ref{ass:D} and \ref{ass:G}, the loss function $\ell^{\expn}_{t,\hat{\alpha}} (\cdot)$ in \eqref{eq:effUSC:surr:exp} is $\frac{\hat{\beta}}{4}$-exp-concave, and $\Vert \nabla \ell_{t,\hat{\alpha}}^{\expn} (\y) \Vert^2  \leq 2G^2$.     
 \end{lem}
 Thus, we use ONS to minimize $\ell^{\expc}_{t,\hat{\alpha}} (\cdot)$. Different from OGD, the projection of ONS onto $\Y$ cannot be achieved through a simple rescaling like Step~5 in Algorithm~\ref{alg:ogd}. Here, we employ an efficient implementation of ONS \citep{pmlr-v99-mhammedi19a} that enhances the efficiency of its projection onto $\Y$. The procedure is summarized in Algorithm~\ref{alg:ons}. 
 \begin{algorithm}[H]
   \caption{Expert $E^i$: ONS for Exp-concave (and Smooth) Functions}\label{alg:ons}
\begin{algorithmic}[1]
   \STATE Let $\y^i_1$ be any point in $\Y$ and $\Sigma_1 =\frac{1}{\hat{\beta}^2D^2} \mathbf{I}_d$
   \FOR{$t=1$ \TO $T$} 
   \STATE Submit $\y^i_t$ to the meta-algorithm
   \STATE Update 
   \begin{equation*}
       \Sigma_{t+1} = \Sigma_t + \nabla \ell^{\expc}_{t,\hat{\alpha}} (\y_t^i) \nabla \ell^{\expc}_{t,\hat{\alpha}} (\y_t^i)^\top,\quad 
       \hat{\y}_{t+1}^i = \y_t^i - \frac{1}{\hat{\beta}} \Sigma_{t+1}^{-1}\nabla \ell^{\expc}_{t,\hat{\alpha}} (\y_t^i)
   \end{equation*}
   where 
   \begin{equation*}
       \nabla \ell^{\expc}_{t,\hat{\alpha}} (\y_t^i) = \nabla g_t(\y_t) + \hat{\beta} \nabla g_t(\y_t)\nabla g_t(\y_t)^\top (\y_t^i-\y_t)
   \end{equation*}
   \STATE Conduct a projection onto $\Y$ 
   \begin{equation*}
    \y^i_{t+1}   = \left\{
\begin{array}{ll}
\hat{\y}^i_{t+1}, & \text{if } \Vert  \hat{\y}^i_{t+1}\Vert\leq D, \vspace{1mm}\\
\mathbf{Q}_{t+1}^\top (4\hat{\beta}D^2 \mathbf{I}_d + \mathbf{\Lambda}_{t+1})^{-1} \mathbf{Q}_{t+1} \Sigma_{t+1}\hat{\y}_{t+1}^i, & \text{otherwise }.
\end{array} \right.
   \end{equation*}
 where $\mathbf{Q}_{t+1}$ and $\mathbf{\Lambda}_{t+1}$ are the matrices of eigenvectors and eigenvalues of $\Sigma_{t+1} - \frac{1}{\hat{\beta}^2D^2} \mathbf{I}_d$
   \ENDFOR
\end{algorithmic}
\end{algorithm}

\subsection{Scale-free Online Gradient Descent for Convex and Smooth Functions}
To exploit smoothness, we use scale-free online gradient descent (SOGD) \citep{Adaptive:Regret:Smooth:ICML} to minimize $\ell_t^{\con}(\cdot)$ in \eqref{eq:effUSC:surr:con}. The procedure is summarized in Algorithm~\ref{alg:sogd}. 
\begin{algorithm}[H]
   \caption{Expert $E^i$: Scale-free OGD for Convex and Smooth Functions}\label{alg:sogd}
\begin{algorithmic}[1]
   \STATE Let $\y^i_1$ be any point in $\Y$
   \FOR{$t=1$ \TO $T$} 
   \STATE Submit $\y^i_t$ to the meta-algorithm
   \STATE Update 
   \begin{equation*}
       \hat{\y}^i_{t+1} = \y^i_t - \eta_t \nabla g_t(\y_t)
   \end{equation*}
   where
   \begin{equation*}
       \eta_t = \frac{\alpha}{\sqrt{\delta+\sum_{s=1}^t \Vert\nabla g_s(\y_s)\Vert^2}},\quad \alpha,\delta >0 
   \end{equation*}
   \STATE Conduct a projection onto $\Y$ 
   \begin{equation*}
    \y^i_{t+1}   = \left\{
\begin{array}{ll}
\hat{\y}^i_{t+1}, & \text{if } \Vert  \hat{\y}^i_{t+1}\Vert\leq D,\vspace{1mm}\\
\hat{\y}^i_{t+1}\cdot \frac{D}{\Vert  \hat{\y}^i_{t+1}\Vert},& \text{otherwise }.
\end{array} \right.
   \end{equation*}
   \ENDFOR
\end{algorithmic}
\end{algorithm}

\subsection{Smooth and Strongly Convex Online Gradient Descent}
\begin{algorithm}[H]
   \caption{Expert $E^i$: Smooth and Strongly Convex OGD}\label{alg:s2ogd}
\begin{algorithmic}[1]
   \STATE Let $\y^i_1$ be any point in $\Y$
   \FOR{$t=1$ \TO $T$} 
   \STATE Submit $\y^i_t$ to the meta-algorithm
   \STATE Update 
   \begin{equation*}
       \hat{\y}^i_{t+1} = \y^i_t - \eta_t \nabla g_t(\y_t)
   \end{equation*}
   where
   \begin{equation*}
       \eta_t = \frac{\alpha}{\delta+\sum_{s=1}^t \Vert\nabla \hat{\ell}^{\strc}_{s,\hat{\lambda}} (\y^i_s)\Vert^2},\quad \alpha,\delta >0 
   \end{equation*}
   \STATE Conduct a projection onto $\Y$ 
   \begin{equation*}
    \y^i_{t+1}   = \left\{
\begin{array}{ll}
\hat{\y}^i_{t+1}, & \text{if } \Vert  \hat{\y}^i_{t+1}\Vert\leq D,\vspace{1mm}\\
\hat{\y}^i_{t+1}\cdot \frac{D}{\Vert  \hat{\y}^i_{t+1}\Vert},& \text{otherwise }.
\end{array} \right.
   \end{equation*}
   \ENDFOR
\end{algorithmic}
\end{algorithm} 
Recall that to exploit smoothness, we enhance the expert-loss for strongly convex functions as follows
\begin{equation*}
     \hat{\ell}^{\strc}_{t,\hat{\lambda}} (\y)= \langle \nabla g_t(\y_t),\y-\y_t\rangle + \frac{\hat{\lambda}}{2G^2} \Vert  \nabla g_t(\y_t) \Vert^2\Vert \y-\x_t\Vert^2.  
 \end{equation*}
 The above expert-loss enjoys the following property. 
 \begin{lem}\label{lem:surrstr:small-loss}
     Under Assumptions~\ref{ass:D} and \ref{ass:G}, $\hat{\ell}^{\strn}_{t,\hat{\lambda}} (\cdot)$ in \eqref{eq:effUSC:surr:str:small-loss} is $\frac{\hat{\lambda}}{G^2}\Vert\nabla g_t(\y_t)\Vert^2$-strongly convex, and $\Vert \hat{\ell}^{\strn}_{t,\hat{\lambda}} (\y)\Vert^2\leq \left( 1+\frac{2D}{G} \right)^2 \Vert \nabla g_t(\y_t)\Vert^2$. 
 \end{lem}
 Due to the modulus of strong convexity is not fixed, we choose Smooth and Strongly Convex OGD (S$^2$OGD) as the expert-algorithm \citep{AAAI:2020:Wang} to minimize $\hat{\ell}^{\strc}_{t,\hat{\lambda}} (\cdot)$. 
 The procedure is summarized in Algorithm~\ref{alg:s2ogd}.

\section{Supporting Lemmas}\label{app:supporting}
\subsection{Proof of Lemma~\ref{lem:surrexp}}
According to the definition of $\ell^{\expc}_{t,\hat{\alpha}} (\cdot)$ in \eqref{eq:effUSC:surr:exp}, we have $\nabla \ell^{\expc}_{t,\hat{\alpha}} (\y) = \nabla g_t(\y_t) + \hat{\beta} \nabla g_t(\y_t) \nabla g_t(\y_t)^\top(\y-\y_t)$. 
Thus, for all $\y\in\Y$, it holds that 
\begin{equation*}
    \begin{aligned}
        \nabla \ell^{\expc}_{t,\hat{\alpha}} (\y) \nabla \ell^{\expc}_{t,\hat{\alpha}} (\y)^\top 
        = {} & \nabla g_t(\y_t) \nabla g_t(\y_t)^\top + 2 \hat{\beta} \nabla g_t(\y_t)(\y-\y_t)^\top \nabla g_t(\y_t)\nabla g_t(\y_t)^\top \\
        &+ \hat{\beta}^2 \nabla g_t(\y_t) \nabla g_t(\y_t)^\top (\y-\y_t) (\y-\y_t)^\top \nabla g_t(\y_t) \nabla g_t(\y_t)^\top \\
        = {} & \left(1+ \hat{\beta} \langle \nabla g_t(\y_t),\y-\y_t  \rangle \right)^2 \nabla g_t(\y_t) \nabla g_t(\y_t)^\top \\
        \preceq {} & 4\nabla g_t(\y_t) \nabla g_t(\y_t)^\top = \frac{4}{\hat{\beta}} \nabla^2 \ell^{\expc}_{t,\hat{\alpha}} (\y)
    \end{aligned}
\end{equation*}
where $\nabla^2 \ell^{\expc}_{t,\hat{\alpha}} (\y)$ denotes the Hessian matrix of $\ell^{\expc}_{t,\hat{\alpha}} (\y)$ and the last inequality is due to a, and the definition of $\hat{\beta}$. Therefore, $\ell^{\expc}_{t,\hat{\alpha}} (\cdot)$ is $\frac{\hat{\beta}}{4}$-exp-concave \citep[Lemma~4.1]{Intro:Online:Convex}. Next, we provide the upper bound of the gradient of $\ell^{\expc}_{t,\hat{\alpha}} (\cdot)$ as follows
\begin{equation*}
    \Vert \nabla \ell^{\expc}_{t,\hat{\alpha}} (\y) \Vert^2\overset{\eqref{eqn:effsurr:G}}{\leq} (G+2\hat{\beta} G^2D)^2 \leq \frac{25}{16}G^2 \leq 2G^2. 
\end{equation*}
This ends the proof. 

\subsection{Proof of Lemma~\ref{lem:surrstr}}
 According to the definition of $\ell_{t}^{\strc} (\cdot)$ in \eqref{eq:effsurr:str}, it holds for any $\x,\y\in \X$ that
 \begin{equation*}
     \ell_{t}^{\strc} (\x) \geq \ell_{t}^{\strc} (\y) + \langle \nabla \ell_{t}^{\strc} (\y),\x-\y \rangle + \frac{\lambda}{2}  \Vert \x -\y \Vert^2. 
 \end{equation*}
 By Definition~\ref{def:strong}, it can be seen that $\ell_{t}^{\strc} (\cdot)$ is $\lambda$-strongly convex. Next, we provide the upper bound of the gradient of $\ell_{t}^{\strc} (\cdot)$ as follows
 \begin{equation*}
     \Vert \nabla \ell_{t}^{\strc} (\y) \Vert^2 \leq \left\Vert \nabla g_t(\y_t)  +\lambda  (\y-\x_t)  \right\Vert^2 
     \overset{\eqref{eqn:effsurr:G}}{\leq} (G+2\lambda D)^2 \leq (G+2 D)^2
 \end{equation*}
 where the last step is due to our assumption that $\lambda\in [1/T,1]$. 

 \subsection{Proof of Lemma~\ref{lem:surrstr:small-loss}}
 Similar to analysis of Lemma~\ref{lem:surrstr}, for any $\x,\y\in \X$, we have
 \begin{equation*}
     \ell_{t,\hat{\lambda}}^{\strc} (\x) \geq \ell_{t,\hat{\lambda}}^{\strc} (\y) + \langle \nabla \ell_{t,\hat{\lambda}}^{\strc} (\y),\x-\y \rangle + \frac{\hat{\lambda}}{2G^2} \Vert\nabla g_t(\y_t)\Vert^2  \Vert \x -\y \Vert^2
 \end{equation*}
 By Definition~\ref{def:strong}, it is established that $\ell_{t,\hat{\lambda}}^{\strc} (\cdot)$ is $\frac{\hat{\lambda}}{G^2}\Vert\nabla g_t(\y_t)\Vert^2$-strongly convex. Next, we upper bound the gradient of $\ell_{t,\hat{\lambda}}^{\strc} (\cdot)$ as follows
 \begin{equation*}
 \begin{aligned}
     \Vert \ell_{t,\hat{\lambda}}^{\strc} (\y)\Vert^2 &\leq \left\langle \nabla g_t(\y_t)+\frac{\hat{\lambda}}{G^2} \Vert\nabla g_t(\y_t)\Vert^2 (\y-\x_t), \nabla g_t(\y_t)+\frac{\hat{\lambda}}{G^2} \Vert\nabla g_t(\y_t)\Vert^2 (\y-\x_t) \right\rangle \\
     & = \Vert \nabla g_t(\y_t)\Vert^2+ \frac{2\hat{\lambda}}{G^2} \Vert\nabla g_t(\y_t)\Vert^2 \langle \nabla g_t(\y_t), \y-\x_t \rangle + \frac{\hat{\lambda}^2}{G^4} \Vert\nabla g_t(\y_t)\Vert^4 \Vert \y-\x_t\Vert^2 \\
     &\overset{\eqref{eqn:effsurr:G}}{\leq}  \left( 1+\frac{2\hat{\lambda}D}{G} \right)^2 \Vert \nabla g_t(\y_t)\Vert^2 \leq \left( 1+\frac{2D}{G} \right)^2 \Vert \nabla g_t(\y_t)\Vert^2
 \end{aligned}
 \end{equation*}
 where the last step is due to our assumption that $\hat{\lambda}\in [1/T,1]$. 

 \subsection{Proof of Lemma~\ref{lem:S2OGD}}
 The analysis is similar to \citet{AAAI:2020:Wang}. Let $\tilde{\y}^i_{t+1} = \y^i_{t}-\frac{1}{\eta_t} \nabla \ell_{t,\hat{\alpha}}^{\strc} (\y^i_{t})$. According to the definition of \eqref{eq:effUSC:surr:str:small-loss}, we have
 \begin{equation*}
 \begin{aligned}
     \ell_{t,k}^{\strc} (\y_t^i) - \ell_{t,k}^{\strc} (\x) &\leq \langle  \nabla \ell_{t,k}^{\strc} (\y^i_t), \y^i_{t} - \x \rangle - \frac{\hat{\lambda}}{2G^2} \Vert \nabla g_t(\y_t) \Vert^2 \Vert \y^i_{t} -\x \Vert^2 \\
     &= \eta_t \langle  \y^i_t -\tilde{\y}^i_{t+1}, \y^i_{t} - \x \rangle- \frac{\hat{\lambda}}{2G^2} \Vert \nabla g_t(\y_t) \Vert^2 \Vert \y^i_{t} -\x \Vert^2. 
 \end{aligned}
 \end{equation*}
 For the first term, it can be verified that 
 \begin{equation*}
     \begin{aligned}
         &\langle  \y^i_t -\tilde{\y}^i_{t+1}, \y^i_{t} - \x \rangle \\
         ={} & \Vert \y_t^i - \x \Vert^2 + \langle  \x -\tilde{\y}^i_{t+1} , \y_t^i - \x \rangle \\
         ={}& \Vert \y_t^i - \x \Vert^2 - \Vert \tilde{\y}^i_{t+1} - \x \Vert^2 - \langle  \y_t^i -\tilde{\y}^i_{t+1} , \tilde{\y}^i_{t+1} - \x \rangle \\
         ={} & \Vert \y_t^i - \x \Vert^2 - \Vert \tilde{\y}^i_{t+1} - \x \Vert^2 + \Vert \tilde{\y}^i_{t+1}-\y_t^i \Vert^2 + \langle \tilde{\y}^i_{t+1}-\y_t^i, \y_t^i-\x \rangle
     \end{aligned}
 \end{equation*}
 which implies that
 \begin{equation*}
     \langle  \y^i_t -\tilde{\y}^i_{t+1}, \y^i_{t} - \x \rangle = \frac{1}{2} \left( \Vert \y_t^i - \x \Vert^2 - \Vert \tilde{\y}^i_{t+1} - \x \Vert^2 + \Vert \tilde{\y}^i_{t+1}-\y_t^i \Vert^2 \right).
 \end{equation*}
 Thus,
 \begin{equation*}
     \begin{aligned}
         \ell_{t,k}^{\strc} (\y_t^i) - \ell_{t,k}^{\strc} (\w) &\leq \frac{\eta_t}{2} \left( \Vert \y_t^i - \x \Vert^2 - \Vert \tilde{\y}^i_{t+1} - \x \Vert^2 \right) \\
         + {} & \frac{1}{2\eta_t} \Vert \nabla \ell_{t,\hat{\alpha}}^{\strc} (\y_t^i) \Vert^2 - \frac{\hat{\lambda}}{2G^2} \Vert \nabla g_t(\y_t) \Vert^2 \Vert \y^i_{t} -\x \Vert^2. 
     \end{aligned}
 \end{equation*}
 Summing the above bound up over $t=1$ to $T$, we attain
 \begin{equation*}
     \begin{aligned}
         & \sum_{t=1}^{T} \ell_{t,\hat{\alpha}}^{\strc} (\y_t^i) - \sum_{t=1}^{T}\ell_{t,\hat{\alpha}}^{\strc} (\x) \\
         \leq {} & \frac{\eta_1}{2} \Vert \y^i_{1} - \x \Vert^2 + \sum_{t=1}^{T} \left( \eta_t - \eta_{t-1} - \frac{\hat{\lambda} }{G^2} \Vert \nabla g_t(\y_t) \Vert^2 \right) \frac{\Vert \y_t^i - \x \Vert^2}{2} 
         + \sum_{t=1}^{T} \frac{1}{2\eta_t}  \Vert \nabla \ell_{t,\hat{\alpha}}^{\strc} (\y_t^i) \Vert^2 \\
         \leq {} & 1+\sum_{t=1}^{T} \frac{1}{2\eta_t}  \Vert \nabla \ell_{t,\hat{\lambda}}^{\strc} (\y_t^i) \Vert^2 \leq 1+ \frac{(G+2D)^2}{2\hat{\lambda}} \sum_{t=1}^{T} \frac{\Vert \nabla g_t(\y_t) \Vert^2}{ (G+2D)^2/\hat{\lambda} + \sum_{i=1}^t \Vert \nabla g_i(\y_i) \Vert^2}.
     \end{aligned}
 \end{equation*}
 where the last two inequalities is due to $\eta_t = (1+2D/G)^2 + \frac{\hat{\lambda}}{G^2} \sum_{i=1}^t \Vert \nabla g_i(\y_i) \Vert^2$ which is specifically set for new expert-loss. Further, we use the following lemma to bound the last term.
 \begin{lem}[{Lemma~11 of \citet{ML:Hazan:2007}}]
     Let $l_1$,$\cdots$,$l_T$ and $\delta$ be non-negative real numbers. Then, we have $\sum_{t=1}^T \frac{l_t^2}{\sum_{i=1}^t l_i^2 + \delta} \leq \log \left( \frac{1}{\delta} \sum_{t=1}^T l_t^2+1\right)$. 
 \end{lem}
 This completes the proof of Lemma~\ref{lem:S2OGD}. 

\section{Clarifications on Bounded Modulus}\label{app:boundmodulus}
In this section, we explain that bounded moduli are generally acceptable in practical scenarios, which is also stated in previous study \citep{ICML:2022:Zhang}. Taking $\lambda$-strongly convex functions as an example, we assume that $\lambda\in [1/T,1]$, since other cases that $\lambda<1/T$ and $\lambda>1$ can be disregarded. 
\begin{itemize}
    \item If $\lambda<1/T$, the regret bound for strongly convex functions becomes $\Omega (T)$, which cannot benefit from strong convexity. Therefore, we should treat these functions as general convex functions. 
    \item If $\lambda>1$, $\lambda$-strongly convex functions are also $1$-strongly convex according to Definition~\ref{def:strong}. Thus, we can treat these functions as $1$-strongly convex functions, and the regret bound is optimal up to a constant factor. 
\end{itemize}

\end{document}